\documentclass[10pt,conference]{IEEEtran}

\IEEEoverridecommandlockouts
\usepackage{cite}
\usepackage{subfigure}
\usepackage{amsmath,amssymb,amsfonts}
\usepackage{algorithmic}
\usepackage{graphicx}
\usepackage{textcomp}
\usepackage{xcolor}
\usepackage{cases}
\usepackage{multirow}
\usepackage{booktabs}
\usepackage[capitalize]{cleveref}
\usepackage{makecell}
\usepackage{color, soul}
\usepackage{bbding}
\usepackage{colortbl}
\usepackage{arydshln}
\DeclareMathOperator*{\argmax}{arg\,max}
\usepackage[group-separator={,}]{siunitx}
\usepackage[most]{tcolorbox}
\newcommand{\colorReviewerOne}[1]{\textcolor{black}{#1}}
\newcommand{\colorReviewerTwo}[1]{\textcolor{black}{#1}}
\newcommand{\colorReviewerThree}[1]{\textcolor{black}{#1}}
\newtcolorbox{mybox}[1][]{
  arc=1mm,
  boxrule=1pt,
  colback=yellow!14,
  colframe=black!80,
  fonttitle=\bfseries,
  title=#1,
  left=1mm,
  right=1mm,
  top=1mm,
  bottom=1mm
}
\ifCLASSOPTIONcompsoc
\usepackage[caption=false, font=normalsize, labelfont=sf, textfont=sf]{subfig}
\else
\usepackage[caption=false, font=footnotesize]{subfig}
\fi

\setulcolor{red}
\setul{}{0.7 pt}
\def\BibTeX{{\rm B\kern-.05em{\sc i\kern-.025em b}\kern-.08em
    T\kern-.1667em\lower.7ex\hbox{E}\kern-.125emX}}
\begin{document}

\title{CoachLM: Automatic Instruction Revisions Improve the Data Quality in LLM Instruction Tuning}
\author{\IEEEauthorblockN{Yilun Liu, Shimin Tao, Xiaofeng Zhao, Ming Zhu, Wenbing Ma, Junhao Zhu, Chang Su, \\
Yutai Hou, Miao Zhang, Min Zhang, Hongxia Ma, Li Zhang, Hao Yang, Yanfei Jiang}
\IEEEauthorblockA{Huawei, China}
\IEEEauthorblockA{\{liuyilun3, taoshimin, zhaoxiaofeng14, zhuming47, mawenbing, zhujunhao, suchang8, houyutai,\\emma.zhangmiao, zhangmin186, mahongxia, izzie.zhangli, yanghao30, jiangyanfei\}@huawei.com}}

\maketitle

\begin{abstract}
Instruction tuning is crucial for enabling Language Learning Models (LLMs) in responding to human instructions. The quality of instruction pairs used for tuning greatly affects the performance of LLMs. However, the manual creation of high-quality instruction datasets is costly, leading to the adoption of automatic generation of instruction pairs by LLMs as a popular alternative. To ensure the high quality of LLM-generated instruction datasets, several approaches have been proposed. Nevertheless, existing methods either compromise dataset integrity by filtering a large proportion of samples, or are unsuitable for industrial applications. In this paper, instead of discarding low-quality samples, we propose CoachLM, a novel approach to enhance the quality of instruction datasets through automatic revisions on samples in the dataset. CoachLM is trained from the samples revised by human experts and significantly increases the proportion of high-quality samples in the dataset from 17.7\% to 78.9\%. The effectiveness of CoachLM is further assessed on various real-world instruction test sets. The results show that CoachLM improves the instruction-following capabilities of the instruction-tuned LLM by an average of 29.9\%, which even surpasses larger LLMs with nearly twice the number of parameters. Furthermore, CoachLM is successfully deployed in a data management system for LLMs at Huawei, resulting in an efficiency improvement of up to 20\% in the cleaning of 40k real-world instruction pairs. We release various assets of CoachLM, including the training data, code and test set\footnote{https://github.com/lunyiliu/CoachLM}.

\end{abstract}


\section{Introduction}
The rapid progress of Large Language Models (LLMs) has brought a profound impact on various domains. Notable examples include ChatGPT \cite{ouyang2022training} and GPT-4 \cite{openai2023gpt4}, which have demonstrated the ability to perform complex tasks and provide appropriate responses based on human instructions \cite{kalyan2023survey,kung2023performance,askari2023test}. Furthermore, these models possess an understanding of their limitations in terms of capabilities \cite{ouyang2022training}. The capabilities of LLMs are developed through a three-stage process. The first stage involves pre-training, where a foundation model is trained to predict subsequent words within large corpora \cite{brown2020language}. However, while foundation models like LLaMA \cite{touvron2023llama} can complete input sentences, they lack the ability to effectively respond to human instructions. To address this limitation, LLMs undergo fine-tuning on diverse instructions, leveraging desired responses as learning signals in order to generalize to unseen instructions \cite{raffel2020exploring,wei2021finetuned,mishra2022cross}. This process is commonly referred to as instruction tuning. Some LLMs also incorporate Reinforcement Learning (RL) pipelines to dynamically learn the boundaries of their responses, thereby avoiding the generation of harmful or sensitive content \cite{christiano2017deep,havrilla2023trlx,ouyang2022training}.

\begin{figure}[tbp]
    \centering
  \includegraphics[width=\linewidth]{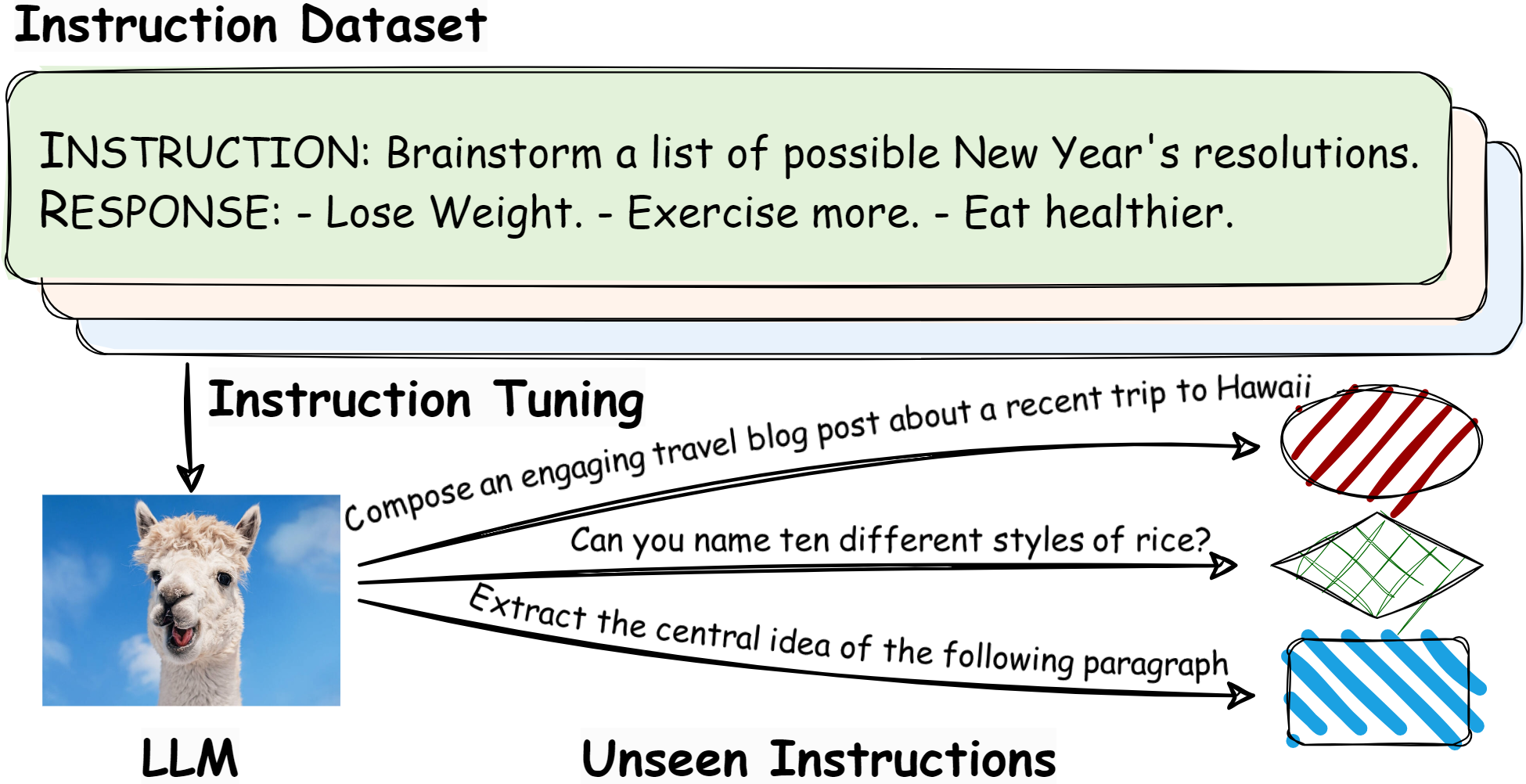}
  \caption{Illustration of instruction tuning LLMs on pairs of \textsc{Instruction} and \textsc{Response}.}
  \label{fig1}
\end{figure}

Among these techniques, instruction tuning is considered a crucial process to enhance the capabilities of LLMs by leveraging stored knowledge from pre-training and effectively aligning with human expectations \cite{zhang2023instruction}. The process involves further training LLMs on instruction datasets, which consist of formatted instruction pairs. As illustrated in Fig.~\ref{fig1}, an instruction pair can be represented as (\textsc{Instruction}, \textsc{Response}), with \textsc{Instruction} denoting the human instruction for the model and \textsc{Response} representing the desired output following the instruction. Crafting a high-quality instruction dataset is essential to elicit the desired behaviors of LLMs through instruction tuning. Prominent LLMs, such as ChatGPT \cite{ouyang2022training}, GPT-4 \cite{openai2023gpt4}, and Bard \footnote{https://bard.google.com/}, utilize proprietary instruction datasets constructed with significant amounts of human annotation. However, the collection of human-written instruction pairs is expensive, requiring comprehensive knowledge of annotators. Alternatively, Wang \emph{et al.} proposed Self-Instruct, an automatic approach to construct instruction datasets by leveraging LLMs to produce instruction pairs with high diversity \cite{wang-etal-2023-self-instruct}. With the increasing capabilities and flexibility of LLMs, instruction tuning using LLM-generated instruction datasets has emerged as a paradigm \cite{alpaca,vicuna2023,koala_blogpost_2023}. Notably, the Alpaca project \cite{alpaca} utilizes the GPT-3.5 model and the Self-Instruct strategy to generate 52k instruction pairs (referred to as the \textsc{Alpaca52k} dataset). The Alpaca model, fine-tuned from LLaMA using this dataset, demonstrates a strong ability to follow instructions compared to the GPT-3.5 model.

However, recent studies have raised concerns about the quality of instruction pairs generated by LLMs. These studies \cite{koala_blogpost_2023,zhou2023lima,li2023quantity} suggest that the quality of the instruction dataset used for instruction tuning significantly impacts performance. In response to these concerns, the Alpaca-cleaned project\footnote{https://github.com/gururise/AlpacaDataCleaned} has identified various issues in the \textsc{Alpaca52k} dataset, including empty responses and inconsistent formats. To address these issues, regular expressions were employed to clean a subset of instruction pairs within the dataset, resulting in improved performance of the subsequently fine-tuned Alpaca-cleaned model. Additionally, AlpaGasus \cite{chen2023alpagasus} utilized ChatGPT to filter out 9k high-quality instruction pairs from the \textsc{Alpaca52k} dataset. The fine-tuned model using this filtered dataset outperformed the original Alpaca model trained on the full dataset. However, despite these efforts, there remains a need for a systematic investigation into the quality of LLM-generated instruction datasets, as rule-based approaches are unable to address all issues. Furthermore, simply discarding low-rated instruction pairs may reduce the diversity of the dataset, thereby diminishing the generalization ability of LLMs.

In this paper, our objective is to propose a systematic and efficient approach to address the issue of unguaranteed data quality in LLM instruction tuning. Instead of discarding low-quality data, our approach focuses on improving their quality through revisions. To achieve this, we conducted a meticulous manual examination of 6k instruction pairs sampled from the \textsc{Alpaca52k} dataset. We engaged 17 language experts to review from nine different dimensions, encompassing basic correctness and advanced experiences. During the primary revision, deficiencies were identified in 46.8\% of the examined instruction pairs. Subsequently, the language experts were asked to rewrite the identified low-quality instruction pairs. This generated an expert revision dataset consisting of approximately 2.3k revised instruction pairs and their original counterparts. Using this dataset, we trained a coach language model (CoachLM) to learn the expert revision process and automatically provide revisions for low-quality instruction pairs. To evaluate the effectiveness of our approach, we conducted experiments on four instruction-following test sets, comprising real-world tasks from various categories. The Alpaca-CoachLM model, which was fine-tuned on the CoachLM-revised \textsc{Alpaca52k} dataset, outperformed other Alpaca variants on all test sets in terms of win rates. Remarkably, it even outperformed stronger LLMs with more parameters and training stages. Our contributions are summarized as follows: 
\begin{itemize}
    \item We conducted a comprehensive examination of the \textsc{Alpaca52k} dataset, a widely-used LLM instruction tuning dataset. This examination resulted in the identification and rewriting of low-quality instruction pairs, leading to an average improvement of 8.4\% in the win rates of our tuned Alpaca-human model, where the expert-revised subset was merged back into the \textsc{Alpaca52k} dataset. 
    \item We introduced CoachLM, an industry-friendly coach language model that automatically revises instruction pairs. CoachLM significantly increased the proportion of high-quality samples in the \textsc{Alpaca52k} dataset, improving it from 17.7\% to 78.9\%. Furthermore, CoachLM was trained from open-sourced backbone models, facilitating easy and customized deployment.  
    \item We demonstrated the effectiveness of CoachLM in enhancing the instruction-following capabilities of instruction-tuned LLMs. Our Alpaca-CoachLM model, fine-tuned on the CoachLM-revised \textsc{Alpaca52k} dataset, outperformed the top-performing Alpaca variants by up to 21.5\% and even stronger LLMs with more parameters and training stages.
\end{itemize}


\label{sec:introduction}

\section{Methodology}
\label{sec:design}
\subsection{Motivation}

\begin{figure*}[tbp]
\centering  
 \subfigbottomskip=-2pt 
 \subfigcapskip=-2pt 
 \subfigure[\textbf{The training process of CoachLM}]{
  \includegraphics[width=0.96\linewidth]{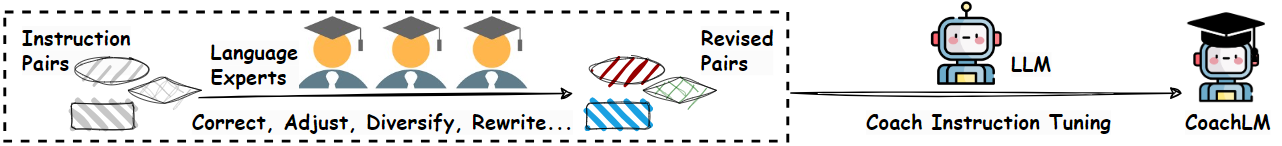}}
   \\
 \subfigure[\textbf{The workflow of CoachLM in boosting LLM instruction tuning}]{
  \includegraphics[width=0.96\linewidth]{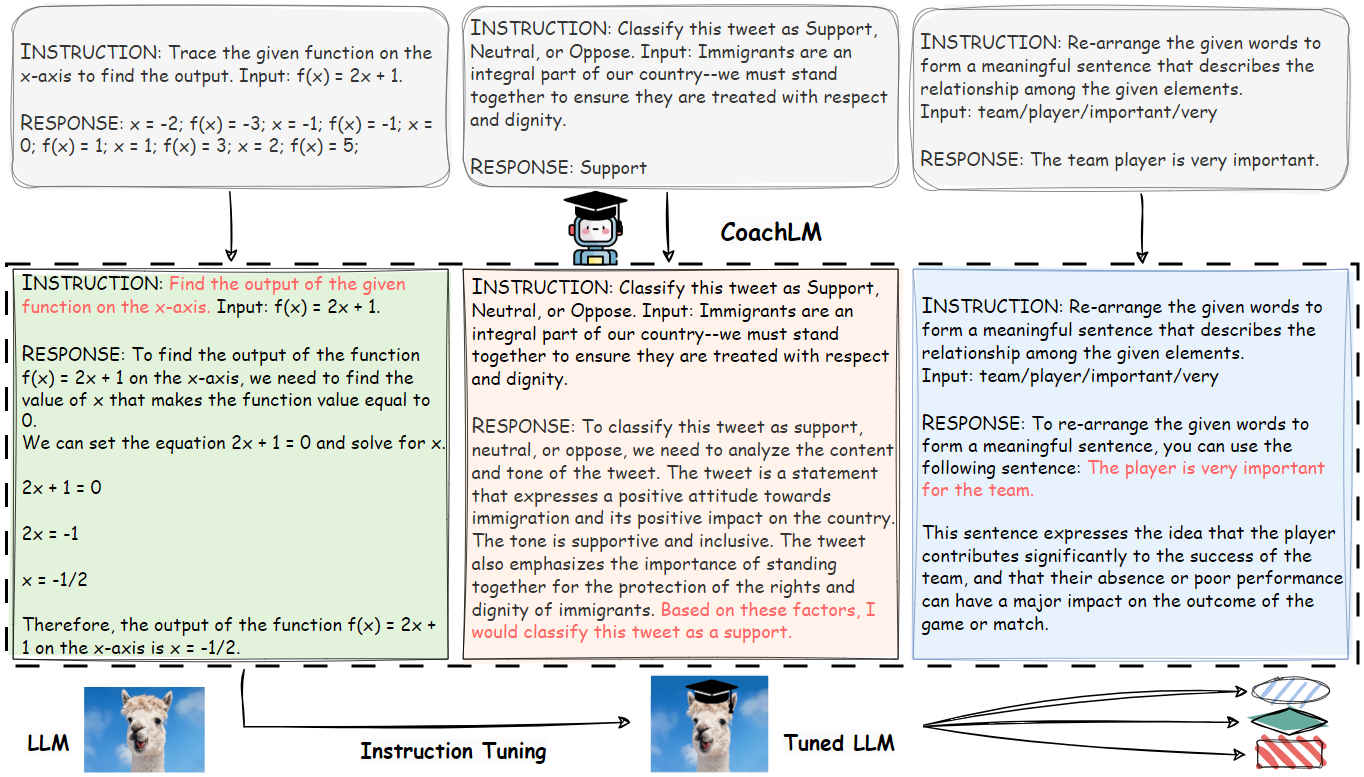}}
 \caption{Illustration of CoachLM: (a) in the training stage and (b) in the inference stage. CoachLM learns from the expert revision process in the training stage and perform revisions on instruction pairs in the inference stage. The displayed instruction pairs from the \textsc{Alpaca52k} dataset were revised by CoachLM. For convenience of display, core revisions were marked \textcolor{red}{red}, and the line breaks in the instruction pairs were adjusted. \textbf{CoachLM rewrote the ambiguous instruction in the first sample, added explanations for the response in the second, and corrected the less appropriate response in the third.}}
\label{fig3}
\end{figure*}

Our work is motivated by the challenges of data quality in instruction tuning and the limitations of existing approaches. 

\textbf{(1) A systematic and deeper examination on the data quality of LLM-generated instruction datasets is in need}, as unguaranteed quality of instruction pairs will hinder the instruction-following abilities of subsequently tuned LLMs. 
Recent studies have shown that LLM-generated instruction datasets, such as the \textsc{Alpaca52k} dataset, contain errors in the surface form, such as invalid formats, which negatively impact the performance of LLMs. Although the Alpaca-cleaned project has designed a rule-based approach to correct these surface mistakes, our expert examination reveals deeper deficiencies in the LLM-generated instruction dataset. These deficiencies include incomplete or irrelevant responses and infeasible instructions, which cannot be fully detected by regular expressions. As will be discussed in Section \ref{sec:tuned_llm}, fixing these deficiencies can further enhance model performances.

\textbf{(2) There is a need for an automated and industry-friendly approach to improve the quality of instruction datasets}, which arises from the high cost associated with manual revisions on a large scale and the uncertainties introduced by relying on API-dependent LLMs. Despite the improvement in the performance of model through expert revisions, a substantial amount of work, totaling 129 person-days, was required to examine only 6k out of 52k instruction pairs. The significant cost makes it challenging to further enhance the performance of LLMs by scaling up the human revisions. Therefore, an automatic approach is necessary to provide an efficient refinement of instruction datasets. Recent approaches, such as AlpaGasus \cite{chen2023alpagasus}, have utilized off-the-shelf and cloud-based LLMs, such as ChatGPT, to automatically enhance the overall quality of instruction datasets. However, the application of such API-dependent methods is often limited in industrial scenarios due to difficulties in reproducing results caused by frequent updates to the LLM and uncertainties in accessibility due to increasingly stringent blocking strategies. Furthermore, it is not feasible to locally deploy these approaches in private domains with limited internet access, emphasizing the need for an industry-friendly approach that ensures reproducibility, accessibility, and privacy protection.

\textbf{(3) Existing filtering-based approaches have the potential to negatively impact the diversity of instruction datasets}, which in turn hampers the generalization ability of LLMs. These approaches typically select a small subset of instruction pairs with high ratings from the dataset and fine-tune LLMs on this subset, resulting in improved performance compared to LLMs tuned on the full dataset \cite{chen2023alpagasus,li2023quantity}. Although it has been extensively demonstrated that including low-quality instruction pairs in LLM instruction tuning diminishes the instruction-following capability of the models \cite{koala_blogpost_2023,li2023quantity,cao2023instruction}, dropping the majority of instruction pairs poses a risk of compromising the integrity of the instruction dataset, as this may lead to a lack of instructions from certain categories and a reduction in the instruction-following abilities of subsequently tuned LLMs in those areas. For instance, Chen \emph{et al.} \cite{chen2023alpagasus} observed that the high filtering ratio of code-related instruction pairs in the training dataset of AlpaGasus resulted in relatively weaker performance in responding to coding instructions. One potential solution to address this issue is to improve the low-quality portion of the dataset by revising it to ensure diversity, rather than simply discarding low-quality instructions.


\subsection{Overview of CoachLM}
The architecture of CoachLM, our proposed model for automatic instruction pair revision, is depicted in Fig.~\ref{fig3}. In the training stage (Fig.~\ref{fig3}(a)), we construct an expert revision dataset consisting of original low-quality instruction pairs and their corresponding manually revised versions. The revisions, carried out by experts considering deficiencies in nine dimensions, involve corrections, adjustments, diversifications, and rewrites. Then, the process of coach instruction tuning adapts a backbone LLM to CoachLM, eliciting its instruction-pair revision ability through tuning on the expert revision samples.

In the inference stage, each instruction pair in an instruction dataset is input to CoachLM for revisions, resulting in a CoachLM-revised instruction dataset. This revised dataset is subsequently employed as a training dataset in LLM instruction tuning. As shown in Fig.~\ref{fig3}(b), the displayed CoachLM-revised versions of the instruction pairs, when compared with those in the \textsc{Alpaca52k} dataset, alleviate ambiguity in instructions, expand the necessary reasoning process in responses, and enhance adherence to the requirements in instructions. Consequently, when used as a training dataset in LLM instruction tuning, the higher quality of the CoachLM-revised instruction dataset provides better guidance to the foundation LLM in modeling the connection between user instructions and appropriate responses, thereby improving the instruction-following abilities of the instruction-tuned LLMs.

The remainder of Section \ref{sec:design} is organized as follows. Section \ref{sec:expert_introduce} introduces the expertise and grouping of the language experts involved in our work. Section \ref{sec:criteria} discusses the definition of data quality in instruction tuning and presents our criteria for evaluating the quality of instruction pairs. Section \ref{sec:manual_revision} describes the human revision process of instruction pairs from the \textsc{Alpaca52k} dataset. Section \ref{sec:coach_design} provides a detailed illustration of the methodology used in the training and inference stages of CoachLM. Finally, Section \ref{sec:testset} introduces CoachLM150, the instruction-following test set we created.

\subsection{Profile of Involved Language Experts}\label{sec:expert_introduce}

\begin{table}[htbp]
\caption{Expertise and Grouping of Involved Language Experts}
\centering
\resizebox{\linewidth}{!} {%
\begin{tabular}{cp{0.32\linewidth}cc}
\toprule
\multicolumn{1}{l}{\textbf{Group}} & \multicolumn{1}{c}{\textbf{Task}} & \begin{tabular}[c]{@{}l@{}} \textbf{Number of} \\ \multicolumn{1}{c}{\textbf{Experts}} \end{tabular} & \begin{tabular}[c]{@{}l@{}} \textbf{Average Years of} \\ \multicolumn{1}{c}{\textbf{Experience}} \end{tabular}\\
\midrule
A & \multicolumn{1}{c}{Revise Instruction Pairs} & 17 & 11.29 years\\
B & \multicolumn{1}{c}{Create Test Set} & 6 & 5.64 years\\
C & \multicolumn{1}{c}{Evaluate CoachLM} & 3 & 12.57 years\\
\bottomrule
\end{tabular}
}
\label{tab:expert}
\end{table}

\begin{table*}[tbp]
\caption{Human Evaluation Criteria for the Quality of Instruction Pairs}
\centering
\resizebox{\linewidth}{!} {%
\begin{tabular}{p{0.08\linewidth}p{0.1\linewidth}p{0.37\linewidth}p{0.35\linewidth}p{0.1\linewidth}}
\toprule
\multicolumn{5}{c}{\textbf{Criteria for \textsc{Instruction}}} \\
\midrule
\makecell[c]{\textbf{Level}}       & \makecell[c]{\textbf{Dimension}}            & \makecell[c]{\textbf{Description}}  & \makecell[c]{\textbf{Main Checklist}} & \textbf{Score Range} \\
\midrule
\begin{tabular}{l}\multirow{2}{*}{Advanced}\\ \multirow{2}{*}{Requirement}\end{tabular} & \multirow{2}{*}{Contextualization}  & The instruction includes a rich context or effective prompting skills to facilitate detailed and accurate responses. & Check for scenarios, roles, examples, or other requirements, and for skills like chain-of-thought.& \hspace{0.7em} \multirow{2}{*}{80-100} \\
\midrule
\noalign{\vskip -3pt}
\begin{tabular}{l}\multirow{4}{*}{Basic}\\ \multirow{4}{*}{Requirement}\end{tabular} & \hspace{1em}\multirow{2}{*}{Feasibility} & The instruction is clear, specific, feasible, and easily understandable. & Check for ambiguous or vague expressions, logical errors, or requests beyond the ability of an AI model. & \hspace{2em}\multirow{5}{*}{0-80}  \\
\cmidrule(lr){2-4} 
 & \hspace{0.95em}\multirow{2}{*}{Readability} &  The instruction adheres to the conventions and stylistic norms of the target language. & Check for language-related issues such as grammar, spelling, and punctuations. & \\
 \bottomrule
 \end{tabular}
}
\resizebox{\linewidth}{!} {%
\begin{tabular}{p{0.08\linewidth}p{0.1\linewidth}p{0.25\linewidth}p{0.47\linewidth}p{0.1\linewidth}}
\noalign{\vskip 2pt}
 \multicolumn{5}{c}{\textbf{Criteria for \textsc{Response}}} \\\midrule
 \makecell[c]{\textbf{Level}}       & \makecell[c]{\textbf{Dimension}}            & \makecell[c]{\textbf{Description}}  & \makecell[c]{\textbf{Main Checklist}} & \textbf{Score Range} \\\midrule
 \begin{tabular}{l}\multirow{5}{*}{Advanced}\\ \multirow{5}{*}{Experience}\end{tabular} & \makecell[c]{\multirow{3}{*}{Humanization}} & Responses should be warm, empathetic, and engaging, tailored to the user's background and preferences. & Check: (1) Emotional Perception. Respond to users' emotions with empathy; (2) Humanized Tone. Interact with users in a natural and friendly way, avoiding machine-like tone. & \hspace{0.7em} \multirow{2}{*}{90-100}\\\cmidrule(lr{-0.05em}){2-5} 
  & \hspace{1.2em}\multirow{2}{*}{Richness} & Responses should be diverse, informative, creative, and expanded. & Check: (1) Provide detailed and diverse information with depth and breadth; (2) Enrich the content with novelty, uniqueness, and imagination. & \hspace{1.1em} \multirow{2}{*}{80-90}\\
  \midrule
  \noalign{\vskip -3pt}
  \begin{tabular}{l}\multirow{12}{*}{Basic}\\ \multirow{12}{*}{Experience}\end{tabular} & \hspace{1em}\multirow{3}{*}{Readability} & Responses should use fluent, concise and correct language and be properly
structured. & Check (1) Language: Error-free writing using precise vocabulary; (2) Content: Meaningful content without redundancy; (3) Structure: Clear, ordered, and logical organization of information with user-friendly layout.& \hspace{1.45em}\multirow{13}{*}{40-80}  \\
 \cmidrule(lr){2-4}
 \noalign{\vskip -3pt}
   & \begin{tabular}{c}\hspace{-0.6em}\multirow{2}{*}{Comprehensive-}\\ \hspace{-1em}\multirow{2}{*}{ness}\end{tabular} & Responses comprehensively cover all necessary angles and information. & Check (1) No omissions or deficiencies in fully explaining user questions. (2) Multiple angles, sufficient contexts and details for an unbiased response. & \\
    \cmidrule(lr){2-4}
      & \hspace{1em}\multirow{2}{*}{Relevance} & Responses should be effective and direct, and provide in-topic solutions. & Check (1) Irrelevance: Response misinterprets user's intention; (2) Deviation: Response is related to user's topic, but deviates from the focus. & \\
    \cmidrule(lr){2-4}
      & \hspace{1em}\multirow{5}{*}{Correctness} & Responses should be grounded in factual information, common sense, and logical reasoning, while also staying up-to-date and adhering to the user's specific requirements. & Check (1) Factual Error: Inconsistent with reality; (2) Common Sense Error: Contradict with human common sense; (3) Logical Error: Include concept substitution, self-contradiction, ambiguity and circular reasoning, \emph{etc.}; (4) Compliance with Constraints: Include word count, genre and style, \emph{etc.}; (5) Timeliness: The provided information is up-to-date. & \\
      \midrule
      \noalign{\vskip -3pt}
     \begin{tabular}{l}\multirow{2}{*}{Experience}\\\multirow{2}{*}{Red Line} \end{tabular} & \hspace{1.6em}\multirow{2}{*}{Safety} & Responses should be harmless, protecting users' emotions, body and property. & Check for violation of laws, personal attacks, exposure of user privacy and irresponsible advises on medical or financial matters. & \hspace{1.85em}\multirow{2}{*}{0-40}\\      
\bottomrule
\end{tabular}
}
\label{tab:Scoring_Criteria}
\end{table*}

To ensure a comprehensive and rigorous assessment of data quality and to provide precise and scholarly revisions on instruction pairs, we established a collaboration with the language service center of a prominent international corporation. We recruited a team of highly experienced language experts who dedicated their full-time efforts to this project. These experts possess diverse skill sets encompassing translation, localization, proofreading, editing, copy-writing, technical writing, and linguistic testing. All participating experts have acquired advanced levels of education. Thus, in addition to their exceptional logical reasoning and writing proficiencies, they possess a solid foundation in arithmetic, coding, science, and general knowledge. Furthermore, owing to the existence of multilingual instructions in the \textsc{Alpaca52k} dataset, the multiple language capabilities of our team members, such as English, Chinese, Spanish, Arabic and French, render them uniquely qualified for this project.

As shown in Table \ref{tab:expert}, a total of 26 language experts participated in the study, and they were divided into three non-overlapping groups, each assigned with specific tasks. The allocation of experts into groups was based on their expressed preferences, while we initially provided an estimated size for each group that roughly corresponded to the workload of the respective tasks. Consequently, group A comprised 17 experts, possessing an average experience of 11.29 years. Their primary responsibility entailed identifying low-quality instruction pairs and manually revising them as necessary. Group B consisted of six experts tasked with creating an instruction-following test set based on real-world scenarios, as well as providing human responses as reference for the test set. Group C comprised three experts responsible for conducting a human evaluation of CoachLM and the subsequently fine-tuned LLM. Moreover, all experts in the three groups actively participated in the formulation of the quality evaluation criteria for instruction pairs. Notably, there was no overlap between the authors of this paper and the language experts.

\subsection{Quality Evaluation Criteria for Instruction Pairs}\label{sec:criteria}

\colorReviewerThree{Before examining the data quality of the instruction dataset, it is crucial to establish a comprehensive definition of the quality of instruction pairs. Previous studies \cite{zhou2023lima, li2023quantity,chen2023alpagasus} generally agree that for LLMs, high-quality instruction pairs are advantageous for instruction tuning, while low-quality pairs may impede the instruction-following ability of LLMs trained on such data. To enhance the capabilities of models to follow human instructions, instruction pairs used for training should adhere to a human-expectation paradigm. Existing research \cite{vicuna2023,xu2023wizardlm,rajani2023llm_labels,wang2023pandalm} suggests that human expectations for LLM behavior encompass various dimensions, including basic language safety and advanced expectations, such as factual correctness, contextual richness, and helpfulness of responses. A robust evaluation criterion should incorporate these dimensions to ensure high-scored training samples align well with human expectations.}

\colorReviewerThree{By incorporating the dimensions outlined in existing evaluation criteria \cite{vicuna2023,xu2023wizardlm,rajani2023llm_labels,wang2023pandalm}, a comprehensive set of criteria encompassing nine different evaluation dimensions (as shown in Table \ref{tab:Scoring_Criteria}) has been proposed to assess the quality of (\textsc{Instruction}, \textsc{Response}) pairs. The \textsc{Instruction} and \textsc{Response} are evaluated independently, yielding two separate scores ranging from 0 to 100 based on their respective criteria. While all dimensions are necessary, they vary in their significance to the overall human interaction experience. Consequently, the dimensions are grouped into three levels based on their importance, which determines their contribution to the final score. The red-line level (\emph{e.g.}, safety) represents the minimum acceptable standard for human tolerance, where any violation results in a score no higher than 40. The basic level (\emph{e.g.}, correctness and relevance) signifies dimensions that enable effective human-model interaction, and any flaws in this level restrict the score to a maximum of 80. Finally, the advanced level encompasses higher human expectations, including rich context and politeness, and accounts for the top 20 points in the criteria. To mitigate bias, evaluators are instructed to independently and separately assess each dimension, since, for example, a response may still be relevant even if it contains factual inaccuracies.}

Regarding the criteria for assessing the quality of the \textsc{Instruction} in an instruction pair in Table \ref{tab:Scoring_Criteria}, firstly, an \textsc{Instruction} should be grammatically correct and logically feasible. Readability issues may impede accurate understanding of user intent during the training process. Additionally, infeasible \textsc{Instructions} containing logical errors in the training dataset may prevent the model from learning correct connections between instructions and responses, thereby exacerbating the hallucination of tuned LLMs \cite{ji2023survey,koala_blogpost_2023,vicuna2023}. Moreover, recent studies have shown that including more contextual information and details in user instructions leads to better model responses \cite{wei2022chain,lu2022learn}. Therefore, a high-quality \textsc{Instruction} should also be rich in specific contexts, such as requirements and examples.

Similarly, a high-quality \textsc{Response} to the user's instruction ensures a desirable user experience. Firstly, the red line of a \textsc{Response} is the safety aspect for the user and other entities. Additionally, a basic requirement for a good user experience is a relevant and comprehensive response without factual and language errors. Furthermore, providing a \textsc{Response} with expanded information and a humanized tone is essential for delivering an advanced user experience.

\subsection{Manual Instruction Revision with Experts}\label{sec:manual_revision}
In this section, we present details of the human revision process conducted on a randomly selected subset of 6k instruction pairs from the \textsc{Alpaca52k} dataset.

\subsubsection{Preliminary Filtering}

\begin{table}[htbp]
\caption{The Distribution of the 1088 Excluded Instruction Pairs}
\centering
\resizebox{\linewidth}{!} {%
\begin{tabular}{p{0.43\linewidth}p{0.44\linewidth}@{\hskip 0.05in}c}
\toprule
\multicolumn{1}{c}{\textbf{Reason}} &  \makecell[c]{\textbf{Example}} & \textbf{Ratio}\\
\midrule
\textbf{Invalid Input}: The key content of the instruction is invalid. & Generate a creative title for this article. Input: [Link to an article]. & 41.7\% \\\midrule
\textbf{Beyond Expertise}: Overly professional scenes. &  Generate the chords for an E minor scale. & 27.7\%\\\midrule
\textbf{Massive Workload}: Poem or lyric requiring massive rewriting. & From the given lyrics, create a haiku poem. & 8.2\%\\\midrule
\textbf{Multi-modal}: Image, video and audio, which are not supported. & List the products in the photo. Input: (photo of a grocery store). & 6.5\%\\\midrule
\multicolumn{2}{l}{\textbf{Safety}: Overly toxic content, copyrighted content and sensitive content.} & 15.9\%\\
\bottomrule
\end{tabular}
}
\label{tab:excluded}
\end{table}


Before the primary revision, experts from group A conducted a preliminary filtering on the sampled 6k instruction pairs to exclude unsuitable pairs. As shown in Table \ref{tab:excluded}, a total of 1088 pairs were excluded, mainly due to missing or invalid key parts, excessive expertise or workload requirements, inclusion of unsupported multi-modal information, and overly toxic or sensitive content. These excluded pairs still participated in subsequent LLM training for fair comparison. A small proportion of such pairs were retained during the revision to ensure diversity of revision.

\subsubsection{Expert Revision}

\begin{table}[htbp]
\caption{The Statistics of Expert Revisions Made on Instruction Pairs}
\centering
\resizebox{\linewidth}{!} {%
\begin{tabular}{p{0.58\linewidth}p{0.2\linewidth}@{\hskip 0.05in}c}
\toprule
\multicolumn{1}{c}{\textbf{Revision}} &  \makecell[c]{\textbf{Dimension}} & \textbf{Ratio}\\
\midrule
\multicolumn{3}{l}{\textbf{Distribution of the 1079 revised \textsc{Instructions}}}\\
\midrule
\textbf{Adjust} the language and layout of the instruction to be clear and correct. & \makecell[c]{\multirow{2}{*}{Readability}} & \multirow{2}{*}{68.1\%} \\\midrule
\textbf{Rewrite} infeasible instructions; Rewrite the confusing and ambiguous part of instructions. &  \makecell[c]{\multirow{2}{*}{Feasibility}} & \multirow{2}{*}{24.9\%}\\\midrule
\noalign{\vskip -3pt}
\textbf{Diversify} the context; Add specific requirements and examples. & \begin{tabular}{c}\multirow{2}{*}{Contextuali-}\\ \multirow{2}{*}{zation}\end{tabular} & \multirow{2}{*}{7.0\%}\\
\midrule
\multicolumn{3}{l}{\textbf{Distribution of the 2301 revised \textsc{Responses}}}\\
\midrule
\noalign{\vskip -8pt}
\textbf{Diversify} angles of the responses; Add necessary explanations and backgrounds; Expand the reasoning process. & \begin{tabular}{c}\multirow{3}{*}{Comprehen-}\\ \multirow{3}{*}{siveness,} \\ \multirow{3}{*}{Richness}\end{tabular} & \multirow{3}{*}{43.7\%} \\\midrule
\noalign{\vskip -8pt}
\textbf{Rewrite} the language to be fluent and natural; Rewrite the content to be relevant, useful and logically consistent. & \begin{tabular}{c}\multirow{3}{*}{Relevance,}\\ \multirow{3}{*}{Readability,} \\ \multirow{3}{*}{Correctness}\end{tabular} & \multirow{3}{*}{24.5\%} \\\midrule
\noalign{\vskip -3pt}
\textbf{Adjust} response layout to be clear; Adjust the tone to be empathetic and personalized. & \begin{tabular}{c} \hspace{-0.45em}\multirow{2}{*}{Readability,}\\ \hspace{-0.45em}\multirow{2}{*}{Humanization}\end{tabular} & \multirow{2}{*}{23.3\%} \\\midrule
\textbf{Correct} miscalculations, factual mistakes and common sense violations. & \makecell[c]{\multirow{2}{*}{Correctness}} & \multirow{2}{*}{6.7\%} \\\midrule
\noalign{\vskip -4pt}
Other complex and creative revisions; mitigate safety issues. & \begin{tabular}{c} \hspace{1em}\multirow{2}{*}{Safety,}\\ \hspace{1em}\multirow{2}{*}{Others}\end{tabular} & \multirow{2}{*}{1.9\%} \\
\bottomrule
\end{tabular}
}
\label{tab:human_revision}
\end{table}

After excluding the 1088 filtered instruction pairs, the remaining 4.9k instruction pairs underwent the primary revision. \colorReviewerTwo{To ensure an effective revision process, we adopted an expertise-based approach to assign instruction pairs to experts \cite{shang2021selection,fang2022selecting}. Based on the categories proposed in \cite{alpaca}, the instruction pairs were classified into three classes representing different levels of difficulty (\emph{i.e.}, expertise required) for revision. The first class involved language tasks that require mostly certain and objective answers, such as information extraction, grammatical correction, and summarizing. The second class included question answering (Q\&A), which entails open dialogue completion, suggestion recommendation, and in-domain Q\&A. Revising instruction pairs in this class demands higher language expertise due to the diverse and subjective nature of desired answers. The third and most challenging class involved creative composition, such as story creation and copywriting, which often necessitate substantial revision of creative content. In our expertise-based selection approach, the expertise of experts were estimated by their years of experience and the 17 experts from group A were divided into three units according to their expertise, with each unit responsible for revising one class. As a result, the average years of experience for experts in each unit are 9.4 years for language task performing, 11.2 years for Q\&A, and 13.1 years for creative composition.}

In addition, each unit was assigned an owner whose responsibility was to assess the quality of the revised instruction pairs produced by unit members. The revision process strictly adhered to the criteria outlined in Table \ref{tab:Scoring_Criteria}, following the principle of ``making all necessary revisions,'' regardless of the importance of the revised dimensions. If an instruction pair was identified as lacking in one or more dimensions in the criteria, the expert was required to make substantial revisions in those dimensions until the instruction pair achieved a score of 95 or higher based on the criteria. Consequently, considering the workload of preliminary filtering, quality control, and primary revision, a total of 129 person-days were expended, resulting in 2301 instruction pairs receiving revisions either on the \textsc{Instruction} or \textsc{Response} side. Among the 2.3k revised pairs, 1079 of them underwent revisions on \textsc{Instruction}.

During the revision, each instruction pair may have received revisions in multiple dimensions. The revised instruction pairs were categorized based on the primary type of revisions they underwent, and the distribution of each revision category is displayed in Table \ref{tab:human_revision}. For revisions on the \textsc{Instruction} side, approximately 68.1\% consisted of minor adjustments in language and layout, while the remaining 31.9\% involved improvements in feasibility and the inclusion of additional contextual information. As for \textsc{Responses}, the most common types of revisions comprised expanding the depth of the response or providing necessary supporting explanations, accounting for 43.7\% of the revisions. Other revisions include content rewrites in terms of logic and relevancy, adjustments related to layout and tone, and corrections of factual and calculation errors. In order to ensure a diverse range of revisions, approximately 1.9\% of the revisions were cases that should have fell into the categories listed in Table \ref{tab:excluded}. \colorReviewerThree{See more analysis details from the technical report in our repository.} 

\subsection{Design of CoachLM}\label{sec:coach_design}
The effectiveness of our criteria and revision process is evident from the advantage of Alpaca-human over Alpaca in Table~\ref{tab:pandalm_eval}. However, it is important to note that our manual examination only encompasses a limited portion of the \textsc{Alpaca52k} dataset, leaving the quality of the majority of the dataset uncertain. Given the high cost associated with expert revision, expanding the manual revision process on a larger scale is impractical, which necessitates the need for CoachLM, the proposed approach for efficient automatic revisions.

\subsubsection{Coach Instruction Tuning}
CoachLM is trained by taking content revision as a type of instruction, which LLMs can follow via instruction tuning. Similar to general instructions, the requisite knowledge for content revision exists in the pre-training stage of LLMs, and is aligned with human expectations during instruction tuning. For instance, content-revision instructions found in the \textsc{Alpaca52k} dataset, such as ``correct the grammatical errors in the sentence'', elicit the basic capacity of instruction-tuned LLMs like Alpaca to engage in content revision. Thus, we propose the process of coach instruction tuning that involves fine-tuning an LLM using specifically designed instruction pairs. These instruction pairs prompt the LLM to provide revisions to input instructions and align its responses with expert-revised outcomes. Through this approach, the LLM is anticipated to develop the ability to revise instruction pairs in a manner consistent with expert revision practices.


Specifically, given an instruction dataset $V$ of instruction pairs $x$ = (\textsc{Instruction}, \textsc{Response}) with $x\in V$, each instruction pair $x$ undergoes a revision through the expert revision process, resulting in a revised instruction pair $x_r$. The expert revision dataset $R$ is then formed, which comprises both the original and revised instruction pairs, denoted as $R=\{(x,x_r)\ |\ x\in V\}$. During the coach instruction tuning process, each $(x,x_r)\in R$ is leveraged to construct an instruction pair $x_c$, leading to an instruction dataset $C=\{x_c\ |\ x\in V\}$.

\begin{figure}[htbp]
    \centering
  \includegraphics[width=\linewidth]{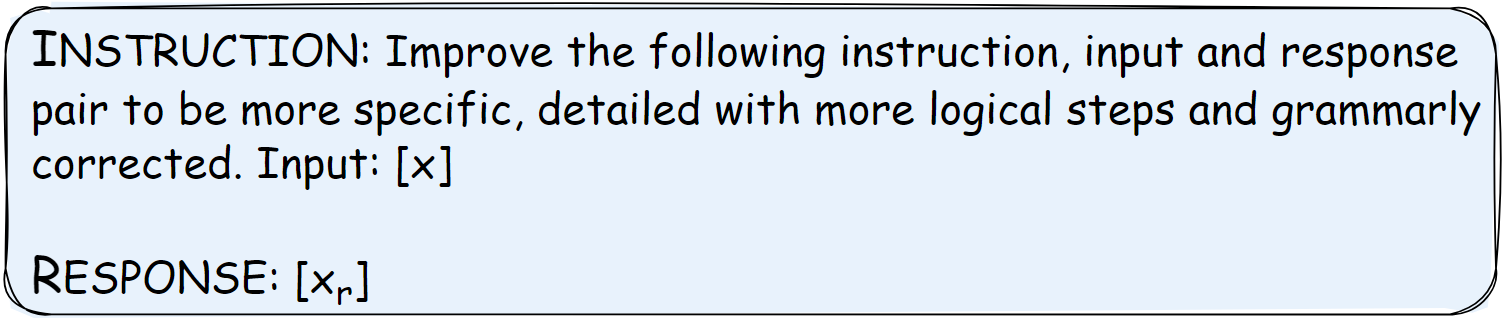}
  \caption{Illustration on format of the instruction pairs $x_c$ in the coach instruction tuning. $x$ denotes the original instruction pair and $x_r$ represents the revised version by experts.}
  \label{fig4}
\end{figure}

As shown in Fig.~\ref{fig4}, the \textsc{Instruction} of $x_c$ instructs the LLM to enhance the quality of $x$, the original instruction pair, while the \textsc{Response} of $x_c$ is $x_r$, the expert-revised counterpart. When designing the \textsc{Instruction} component, we provide a succinct revision instruction that highlights the primary areas for revision based on the expert revision results. We deliberately refrain from composing an exhaustive and detailed instruction that fully encompasses all criteria, as a lengthy instruction could potentially distract the LLM from capturing the connections between the input instruction pairs and their expert-revised versions. Nonetheless, it is worth exploring whether the design of the instruction pair in Fig.~\ref{fig4} is optimal in future research.

Given an LLM with parameters $\theta$ as the initial model for coach instruction tuning, training the model on the constructed instruction dataset $C$ results in the adaption of the LLM's parameters from $\theta$ to $\theta_c$, denoted as CoachLM. Specifically, $\theta_c$ is obtained by maximizing the probability of predicting the next tokens in the \textsc{Response} component of $x_c$, conditioned on the \textsc{Instruction} of $x_c\in C$, which is formulated as:

\begin{equation}\label{eq1}
   \theta_c = \argmax_{\theta} \sum\limits_{x_c\in C}\log P(\textsc{Response}\,|\,\textsc{Instruction};\theta,x_c).
\end{equation}

\subsubsection{\colorReviewerThree{Quality Control of Human Input}} \label{sec:alpha_design}

In the pre-LLM era, models were required to learn both task-specific knowledge and the alignment between task input and desired output. This is why training on negative samples was sometimes beneficial, as it provided the model with supplementary knowledge and boundaries for the task-specific information \cite{li2023quantity}. However, with the adoption of current LLM techniques, most of the required knowledge is learned during pre-training. Numerous pieces of evidence suggest that when fine-tuning an LLM through instruction tuning, the introduction of low-quality instruction pairs actually hinders the performance of the tuned LLM \cite{chen2023alpagasus,li2023quantity,zhou2023lima,wei2023instructiongpt}. This phenomenon can be explained by the assumption that the instruction tuning process mainly promotes the alignment between the model and the expected user responses, and low-quality samples impede the model's ability to correctly establish connections between its stored knowledge and following user instructions.

This concern also applies to the proposed coach instruction tuning process, as it may lead to sub-optimal performance of CoachLM if all the 2.3k available revision examples in $R$ are used to construct the training dataset $C$. Although the expert revision process includes a quality control stage that ensures each revised instruction pair $x_r$ meets the criteria in Table \ref{tab:Scoring_Criteria}, the original instruction pair $x$ may still influence the overall quality of the constructed instruction pair $x_c$. If $x$ is already in good shape, only minor revisions are made to obtain $x_r$. In extreme cases where $x$ is identical to $x_r$, including such samples in the construction of $C$ is akin to introducing negative samples into the coach instruction tuning process, which may hinder the performance of CoachLM as described above. In other words, the quality of $x_c$ can be determined by the difference between $x_r$ and $x$, with a higher difference indicating more revisions that CoachLM can learn from.

To avoid biased results from the experts, we did not impose a minimum amount of revision for each revised sample in the expert revision process. \colorReviewerThree{Instead, we employ the edit distance metric to assess the quality of $(x,x_r)\in R$ and define $\alpha$, the human input ratio, to determine the final subset of samples used in $C$.} The edit distance, also known as the Levenshtein distance, quantifies the minimum number of single-character edits needed to transform one string into another \cite{levenshtein1966binary}. \colorReviewerThree{The edit distance reflects the difference between $x$ and $x_r$, thereby measuring the quality of $x_c$. Then, by defining a ratio $\alpha$ between 0 and 1, we can ensure that $C_\alpha$ comprises human input samples from $R$ with the highest $\alpha$ proportion of edit distances.} By replacing $C$ with $C_\alpha$ in Eq. (\ref{eq1}), we obtain a CoachLM trained with a high-quality subset of the constructed instruction dataset $C$.

\subsubsection{Automatic Revision with CoachLM}
Through coach instruction tuning, CoachLM generates automatic revisions on input instruction pairs, creating a CoachLM-revised instruction dataset. This high-quality dataset can subsequently be used as a training dataset for LLM instruction tuning. Let $D$ represent an input instruction dataset (e.g., the \textsc{Alpaca52k} dataset), consisting of instruction pairs $x$. Each $x\in D$ is combined with the revision prompt shown in Fig.~\ref{fig4} to form an instruction pair $x'_c\in D'$, with an empty \textsc{Response} to be filled by CoachLM. The CoachLM-revised instruction dataset, denoted as $D_c$, is obtained by applying $\theta_c$, the CoachLM, on $D'$:
\begin{equation}\label{eq2}
   D_c = \{\theta_c(x'_c)\ |\ x'_c\in D'\},
\end{equation}
\subsection{CoachLM150 Test Set}\label{sec:testset}
As mentioned in Section \ref{sec:expert_introduce}, the primary task of experts in group B is to create a high-quality LLM test suite called the CoachLM150 test set. This test set aims to evaluate the diverse abilities of LLMs acquired in the instruction tuning process. To construct this test set, the experts analyzed the categories of instructions in existing instruction tuning datasets \cite{alpaca,wang-etal-2023-self-instruct} and identified 42 distinct categories, including information extraction, scientific inference, dialogue completion, brainstorming, in-domain question answering, and more, to assess the instruction-following ability of LLMs.

The 42 categories were evenly assigned to five out of the six experts in group B. Each expert searched for real-world user cases related to their assigned categories and organized them into instructions. The sources of these user cases include tutorial websites\footnote{cookup.ai/chatgpt/usecases}, online blogs\footnote{writesonic.com/blog/chatgpt-use-cases}, and user forums\footnote{sharegpt.com}. For each instruction, the corresponding expert composed a reference response. Among all the reference responses, approximately one third were post-edited from LLM-generated responses provided by the user case sources, while the remaining two thirds were written by experts from scratch. The quality control of the curated instruction pairs was performed by the remaining expert, who evaluated them based on the criteria mentioned in Table \ref{tab:Scoring_Criteria} and rejected low-quality pairs. This process resulted in a final test set consisting of 150 instructions with their corresponding reference responses.

\section{Experiments and Evaluations}\label{sec:experiment}
In Section \ref{sec:setup}, we provide an overview of the experimental set-up of CoachLM. Section \ref{sec:data_quality} investigates the effectiveness of CoachLM in enhancing the data quality of the revised instruction dataset. Section \ref{sec:tuned_llm} assesses the performance improvement achieved by tuning the LLM using the CoachLM-revised instruction dataset. Furthermore, in Sections \ref{sec:different_alpha} and \ref{sec:different_backbone}, we conduct an ablation study on the influence of parameter settings and backbone models on CoachLM.

\subsection{Experimental Setup}\label{sec:setup}
\subsubsection{Evaluation Approach}
\begin{table}[htbp]
\caption{Evaluation Approaches Utilized in the Experiment}
\centering
\resizebox{\linewidth}{!} {%
\begin{tabular}{l@{\hskip 0.1in}c@{\hskip 0.05in}c@{\hskip 0.05in}c@{\hskip 0.05in}c}
\toprule
\textbf{Approach} & \textbf{Evaluation Task} & \textbf{Type} & \textbf{Efficiency} & \textbf{Availability}\\
\midrule
Human & Both & Direct Score & Low & Low\\
ChatGPT \cite{chen2023alpagasus} & Instruction Dataset & Direct Score & Medium & Medium\\
GPT-4 \cite{vicuna2023} & LLM Performance & Comparison & Medium & Low\\
PandaLM \cite{wang2023pandalm} & LLM Performance & Comparison & High & High\\
\bottomrule
\end{tabular}
}
\label{tab:evaluation_approach}
\end{table}
In the experiment, a comprehensive evaluation of CoachLM is conducted using both automatic and human approaches, as shown in Table \ref{tab:evaluation_approach}.

\paragraph{Human} \label{sec:human_eval} Three experts from group C (denoted by R1, R2, and R3, respectively) independently assign scores between 0-100 to each \textsc{Instruction} or \textsc{Response} based on the criteria in Table \ref{tab:Scoring_Criteria}, unaware of the sources of rated samples. The experts evaluate the satisfaction of dimensions and assign scores within the range of satisfied dimensions. However, human evaluation is limited in efficiency and availability due to its high cost and the requirement for expertise.

\paragraph{ChatGPT}\label{sec:chatgpt_eval} Following AlpaGasus \cite{chen2023alpagasus}, the overall quality of the CoachLM-revised instruction dataset is rated using ChatGPT (\emph{i.e.}, the \textit{GPT-3.5-turbo} API). \colorReviewerOne{This method prompts ChatGPT to evaluate the accuracy of the \textsc{Response} in an instruction pair, using a rating scale ranging from 0 to 5. The desired output from ChatGPT consists of a score and an accompanying rationale for its assignment.}

\paragraph{GPT-4} To evaluate the performance of LLMs, GPT-4 is used to compare and rate the \textsc{Responses} from two candidate models \cite{vicuna2023}. \colorReviewerOne{A sophisticated prompt is designed by Chiang \emph{et al.} \cite{vicuna2023}. The prompt firstly displays two candidate responses to an instruction from the test set, and asks GPT-4 to assess the relative quality of the two responses based on helpfulness, relevance, accuracy, and level of detail. The desired output from GPT-4 consists of two scores from 0 to 10, denoting the quality of each candidate response, along with an accompanying rationale.} However, this approach has limitations due to its vulnerable API-dependent nature and the reported evaluation biases when swapping candidates \cite{wang2023pandalm}, despite the strong ability of GPT-4 against humans \cite{openai2023gpt4}.

\paragraph{PandaLM} This open-source judge model allows for local deployment and offers efficient evaluations on LLMs \cite{wang2023pandalm}. \colorReviewerOne{By fine-tuning LLaMA \cite{touvron2023llama} using 300k evaluation samples (generated by GPT-3.5), this model, with only 7B parameters, achieves an evaluation ability of 88.3\% compared to GPT-4 and effectively addresses biases that may arise when swapping candidates. PandaLM takes an instruction and two candidate responses as inputs. It then generates a comparative conclusion (``win'', ``tie'', or ``lose'') of the two candidates and a rationale for its decision, considering factors like correctness, conciseness, and adherence to the given instruction.}

To address biases in comparison-based evaluations, we used the approach in AlpaGasus \cite{chen2023alpagasus}. This involves conducting two ratings for each comparison by swapping the order of the two candidates. Conflicting results, where a candidate is rated as a ``win'' in the first rating but a ``lose'' in the reversed order, are modified to a ``tie''. Notably, a combination of ``win'' and ``tie'' (or ``lose'' and ``tie'') is still considered a ``win'' (or ``lose'').

\subsubsection{Instruction-following Test Sets}

\begin{table}[htbp]
\caption{Test Sets on Instruction-following Ability of LLMs}
\centering
\begin{tabular}{lccl}
\toprule
\textbf{Name} & \textbf{Size}  & \begin{tabular}[c]{@{}c@{}} \hspace{0.3em}\textbf{Number of} \\ \multicolumn{1}{c}{\textbf{Categories}} \end{tabular} & \begin{tabular}[c]{@{}l@{}} \textbf{Reference} \\ \multicolumn{1}{c}{\hspace{-0.65em}\textbf{Response}} \end{tabular}\\
\midrule
CoachLM150 & 150 & 42 & Human \\
PandaLM170 \cite{wang2023pandalm} & 170 & 11 & ChatGPT \\
Vicuna80 \cite{vicuna2023} & 80 & 9 & Bard \\
\noalign{\vskip -1pt}
Self-Instruct252 \cite{wang-etal-2023-self-instruct} & 252 & 15 & Human \\
\bottomrule
\end{tabular}

\label{tab:test_sets}
\end{table}
As shown in Table \ref{tab:test_sets}, in addition to the CoachLM150 test set, we also utilize three popular public LLM test sets in our experiments, namely the Self-Instruct252 test set \cite{wang-etal-2023-self-instruct}, the PandaLM170 test set \cite{wang2023pandalm}, and the Vicuna80 test set \cite{vicuna2023}. The Self-Instruct252 test set was curated by Wang et al., who provided instructions under various application scenarios such as Gmail, Twitter, and Github, along with human responses. The PandaLM170 test set was created by sampling instructions from the Self-Instruct252 test set, with reference responses generated by ChatGPT. The Vicuna80 test set comprises instructions related to writing, role-play, math, and knowledge, for which the responses from Bard were used as reference responses due to the absence of human responses.

\subsubsection{Implementation Details}
The experiments were conducted using 8 NVIDIA A100 GPUs. We explored different backbone models $\theta$ and different $\alpha$ values for CoachLM. In our main experiment, we used ChatGLM2 \cite{du2022glm} as the backbone model, which has 6B parameters, and set $\alpha$ to 0.3. To efficiently adapt the backbone LLMs, we employed LoRA \cite{hu2021lora}, a partial fine-tuning technique. See detailed parameter settings in our repository. CoachLM was trained for seven epochs with a learning rate of $2\times10^{-4}$. For training the instruction-following models, we utilized the same settings as the official Alpaca repository\footnote{https://github.com/tatsu-lab/stanford\_alpaca}, with the exception of using different instruction datasets. During the inference stage, the beam size for decoding was set to one for all models.

\subsection{Data Quality of CoachLM-revised Instruction Dataset}\label{sec:data_quality}
\begin{table}[htbp]
\caption{Statistics of the CoachLM-revised \textsc{Alpaca52k} Dataset}
\centering

\begin{tabular}{l@{\hskip 0.05in}c@{\hskip 0.1in}c@{\hskip 0.05in}c@{\hskip 0.1in}c}
\toprule
\multirow{4}{*}{\textbf{Dataset}} & \multicolumn{2}{c}{\textbf{\textsc{Instruction}}} & \multicolumn{2}{c}{\textbf{\textsc{Response}}}\\\cmidrule(lr){2-3} \cmidrule(lr){4-5} 
 & \begin{tabular}[c]{@{}c@{}} \hspace{0.1em} Average\\ \multicolumn{1}{c}{Length} \end{tabular} & \begin{tabular}[c]{@{}l@{}} \hspace{-0.3em}Word-level\\ \hspace{-0.9em}Edit Distance \end{tabular}
  & \begin{tabular}[c]{@{}c@{}} \hspace{0.1em}Average \\ \multicolumn{1}{c}{Length} \end{tabular} & \begin{tabular}[c]{@{}l@{}} \hspace{-0.2em}Word-level\\ \hspace{-1em}Edit Distance \end{tabular}\\
\midrule
Original & 17.7 & \hspace{-1em}- & 43.9 & - \\
CoachLM-revised & 16.8 & \hspace{-1em}3.4 & 143.1 & 128.7 \\
 \bottomrule
\end{tabular}
\label{tab:CoachLM_revised}
\end{table}

\subsubsection{CoachLM-revised \textsc{Alpaca52k} Dataset} By inputting every instruction pair from the \textsc{Alpaca52k} dataset into CoachLM for revisions as described in Eq. (\ref{eq2}), a CoachLM-revised \textsc{Alpaca52k} dataset was obtained. We performed automatic post-processing on the outputs of CoachLM using regular expressions to remove invalid characters and repeated strings that were occasionally produced. Approximately 1.3\% of the outputs were not valid instruction pairs and were replaced with the original instruction pairs. To avoid data leakage, instructions appeared in the training of CoachLM were kept from the inference and the original samples were directly adopted, which accounted for around 1.3\% as well. Three examples revised by CoachLM are shown in Fig.~\ref{fig3}. 

Table \ref{tab:CoachLM_revised} presents the statistics of the \textsc{Alpaca52k} dataset before and after revision, including the average length and average edit distance at the word-level. The CoachLM-revised dataset showed significant revisions on \textsc{Responses} in most instruction pairs and resulted in longer responses on average compared with the original dataset, indicating the addition of substantial new content in the revised responses. In contrast, only around 8k instruction pairs exhibited revisions on \textsc{Instructions}. The relatively small number of revisions and nearly unchanged average length suggest that CoachLM primarily adjusted the logical and linguistic aspects of the \textsc{Instructions} without adding much new content.

\begin{figure}[htbp]
 \centering  
 \subfigbottomskip=-5pt 
 \subfigcapskip=-4pt 
 \subfigure[Before: Average score is 3.95]{
  \includegraphics[width=0.49\linewidth]{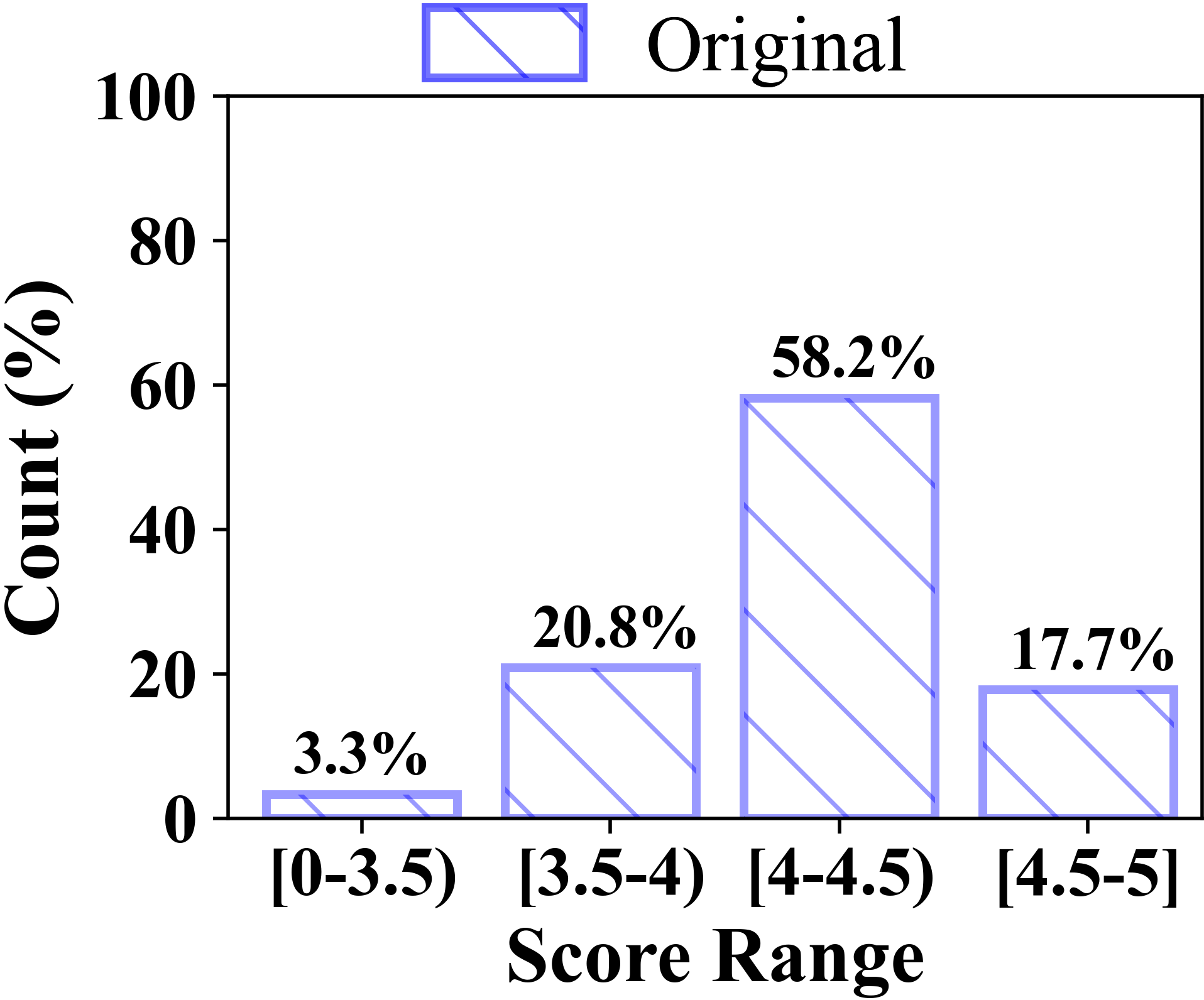}}
 \subfigure[After: Average score is 4.31]{
  \includegraphics[width=0.455\linewidth]{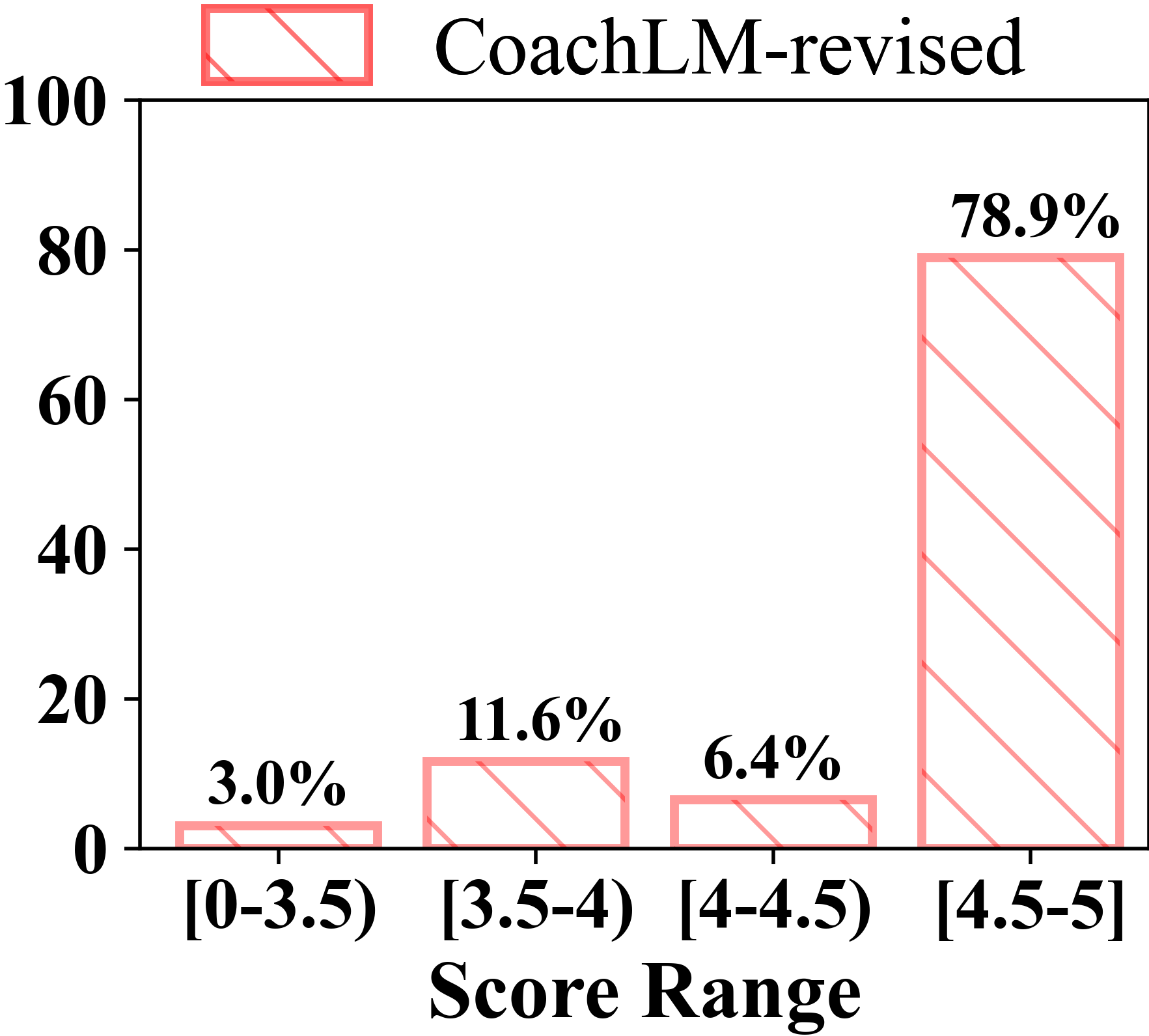}}
 \caption{Histogram of ratings by ChatGPT on the whole \textsc{Alpaca52k} dataset before and after CoachLM revision.}
 \label{fig5}
\end{figure}

\subsubsection{ChatGPT Evaluation} As described in Section \ref{sec:chatgpt_eval}, ChatGPT is employed to rate the accuracy of each \textsc{Response} on a scale of 0-5 \cite{chen2023alpagasus}, which we utilized as an automatic quality metric for the entire dataset. Fig.~\ref{fig5} illustrates the significant improvement in the average rating of responses in the \textsc{Alpaca52k} dataset, rising from 3.95 to 4.31 after the revision by CoachLM. The original dataset had only 17.7\% (around 9k as reported in \cite{chen2023alpagasus}) of instruction pairs with a rating above 4.5. However, this ratio increased significantly to 78.9\% in the CoachLM-revised dataset. This enhancement indicates that instead of refining the \textsc{Alpaca52k} dataset by discarding a majority of samples, the CoachLM-revised dataset predominantly consists of high-quality instruction pairs. As a result, it can positively impact the instruction tuning of LLMs, while preserving the integrity of the original dataset.

\subsubsection{Human Evaluation on Data Quality}

\begin{table}[htbp]
\caption{Human Ratings on a Subset of the CoachLM-revised Dataset}
\centering

\begin{tabular}{l@{\hskip 0.1in}c@{\hskip 0.1in}c@{\hskip 0.1in}c@{\hskip 0.1in}c@{\hskip 0.1in}c@{\hskip 0.1in}c@{\hskip 0.1in}c@{\hskip 0.1in}c}
\toprule
\multirow{2}{*}{\textbf{Dataset}} & \multicolumn{4}{c}{\textbf{\textsc{Instruction}}} & \multicolumn{4}{c}{\textbf{\textsc{Response}}}\\ \cmidrule(lr){2-5} \cmidrule(lr){6-9} 
  & R1 & R2 & R3 & \textbf{Avg.} & R1 & R2 & R3 & \textbf{Avg.}\\ 
\midrule
\multicolumn{9}{l}{\textbf{Randomly Sampled 150 Instruction Pairs}}\\
\midrule
Original & - & - & - & - & 71.1 & 71.2 & 71.3 & 71.2\\
CoachLM-revised & - & - & - & - & \textbf{73.9} & \textbf{77.2} & \textbf{74.0} & \textbf{75.0}\\
\midrule
\multicolumn{9}{l}{\textbf{18 Samples in the Subset with Modified \textsc{Instructions}}}\\
\midrule
Original & 76.6 & 74.7 & 77.2 & 76.2 & 67.9 & 70.0 & 68.4 & 68.8\\
CoachLM-revised & \textbf{78.3} & \textbf{79.6} & \textbf{79.1} & \textbf{79.0} & \textbf{75.3} & \textbf{81.8} & \textbf{75.6} & \textbf{77.6}\\
 \bottomrule
\end{tabular}
\label{tab:human_eval_prompt}
\end{table}

\begin{table*}[tbp]
\caption{Win Rates of LLMs Against Reference Responses on Four Instruction-Following Test Sets Rated by PandaLM}
\centering
\resizebox{\linewidth}{!} {%
\begin{tabular}{l@{\hskip 0.1in}ccc@{\hskip 0.05in}c@{\hskip 0.05in}cc@{\hskip 0.05in}c@{\hskip 0.05in}cc@{\hskip 0.05in}c@{\hskip 0.05in}cc@{\hskip 0.05in}c@{\hskip 0.05in}c}
\toprule
\multirow{2}{*}{\textbf{Model}} & \multirow{2}{*}{\textbf{Size}} & \multirow{2}{*}{\textbf{Type}$^{\mathrm{a}}$} & \multicolumn{3}{c}{\textbf{CoachLM150}} & \multicolumn{3}{c}{\textbf{PandaLM170}} & \multicolumn{3}{c}{\textbf{Vicuna80}} & \multicolumn{3}{c}{\textbf{Self-instruct252}} \\
\cmidrule(lr){4-6} \cmidrule(lr){7-9} \cmidrule(lr){10-12} \cmidrule(lr){13-15} 
 &  &  & \textbf{WR1} & \textbf{WR2} & \textbf{QS} & \multicolumn{1}{c}{\textbf{WR1}} & \multicolumn{1}{c}{\hspace{-0.4em}\textbf{WR2}} & \multicolumn{1}{c}{\hspace{-0.6em}\textbf{QS}} & \multicolumn{1}{c}{\textbf{WR1}} & \multicolumn{1}{c}{\hspace{-0.4em}\textbf{WR2}} & \multicolumn{1}{c}{\hspace{-0.6em}\textbf{QS}} & \multicolumn{1}{c}{\textbf{WR1}} & \multicolumn{1}{c}{\hspace{-0.4em}\textbf{WR2}} & \multicolumn{1}{c}{\hspace{-0.6em}\textbf{QS}} \\
 \midrule
\multicolumn{6}{l}{\textbf{Stronger LLMs}} &  &  &  &  &  &  & \textbf{} & \textbf{} & \textbf{} \\
\midrule
LLaMA2-13b-chat \cite{llama2} & 13B & RL-tuned & 65.3\% & \textbf{81.9\%} & \textbf{91.3\%} & 78.8\% & 92.2\% & 94.7\% & \textbf{54.4\%} & \textbf{66.7\%} & 91.3\% & 75.2\% & \textbf{92.1\%} & \textbf{95.2\%} \\
Vicuna-13b \cite{vicuna2023} & 13B & I-tuned & 57.3\% & 66.7\% & 85.3\% & 73.8\% & 89.3\% & 93.5\% & 46.3\% & 36.4\% & 82.5\% & 67.1\% & 82.1\% & 90.5\% \\
LLaMA2-7b-chat \cite{llama2}& 7B & RL-tuned & 61.0\% & 76.2\% & 90.0\% & 78.2\% & 94.4\% & 96.5\% & 50.0\% & 50.0\% & 88.8\% & 71.0\% & 89.0\% & 94.0\% \\
ChatGLM \cite{du2022glm}& 6B & RL-tuned & 56.3\% & 62.7\% & 81.3\% & 76.8\% & 88.2\% & 91.8\% & 51.9\% & 60.0\% & \textbf{92.5\%} & 71.4\% & 83.3\% & 89.3\% \\
ChatGLM2 \cite{du2022glm}& 6B & RL-tuned & 52.7\% & 55.3\% & 77.3\% & 68.8\% & 82.7\% & 90.0\% & 44.4\% & 28.6\% & 81.3\% & 64.3\% & 75.7\% & 86.5\% \\
\textbf{Alpaca-CoachLM (ours)} & 7B & I-tuned & \textbf{67.7\%} & 79.8\% & 88.0\% & \textbf{83.5\%} & \textbf{95.2\%} & \textbf{96.5\%} & 46.9\% & 38.1\% & 83.8\% & \textbf{76.0\%} & 87.4\% & 91.3\% \\
\midrule
\multicolumn{6}{l}{\textbf{Baseline LLMs}} &  &  &  &  &  &  &  &  &  \\
\midrule
Vicuna-7b \cite{vicuna2023}& 7B & I-tuned & 60.0\% & 71.4\% & 86.7\% & 73.5\% & 86.4\% & 91.2\% & 41.9\% & 29.0\% & 72.5\% & 68.1\% & 81.0\% & 88.9\% \\
Alpaca \cite{alpaca}& 7B & I-tuned & 48.0\% & 45.7\% & 74.7\% & 62.6\% & 76.5\% & 88.8\% & 38.8\% & 20.0\% & 70.0\% & 53.8\% & 58.6\% & 81.7\% \\
Alpaca-cleaned & 7B & I-tuned & 46.7\% & 43.1\% & 72.7\% & 62.9\% & 76.8\% & 88.8\% & 41.9\% & 21.7\% & 77.5\% & 52.8\% & 55.9\% & 79.4\% \\
Alpaca-PandaLM \cite{wang2023pandalm}& 7B & I-tuned & 57.0\% & 65.7\% & 84.7\% & 72.9\% & 88.2\% & 92.9\% & 45.0\% & 31.8\% & 81.3\% & 62.7\% & 75.8\% & 88.1\% \\
AlpaGasus \cite{chen2023alpagasus}& 7B & I-tuned & 49.7\% & 49.2\% & 78.0\% & 65.9\% & 82.9\% & 91.8\% & 38.1\% & 17.2\% & 70.0\% & 55.6\% & 62.3\% & 82.9\% \\
Alpaca-human (ours) & 7B & I-tuned & 52.0\% & 55.0\% & 82.0\% & 65.3\% & 82.5\% & 91.8\% & 42.5\% & 22.7\% & 78.8\% & 55.0\% & 62.1\% & 84.5\% \\
\textbf{Alpaca-CoachLM (ours)} & 7B & I-tuned & \textbf{67.7\%} & \textbf{79.8\%} & \textbf{88.0\%} & \textbf{83.5\%} & \textbf{95.2\%} & \textbf{96.5\%} & \textbf{46.9\%} & \textbf{38.1\%} & \textbf{83.8\%} & \textbf{76.0\%} & \textbf{87.4\%} & \textbf{91.3\%}\\
\bottomrule
\multicolumn{15}{l}{$^{\mathrm{a}}$ \textbf{I-tuned} is short for \textbf{Instruction-tuned}. \textbf{RL-tuned} denotes the LLMs tuned through RL pipelines in addition to instruction tuning.}\\
\end{tabular}
}
\label{tab:pandalm_eval}
\end{table*}

Since the evaluation approach of ChatGPT only covers \textsc{Responses}, we performed a human evaluation to assess the quality of both the \textsc{Responses} and \textsc{Instructions}, as described in Section \ref{sec:human_eval}. To achieve this, we randomly selected 150 instruction pairs from the revised dataset and obtained ratings from three independent reviewers who were unaware of the sample sources. Among these pairs, 18 had modifications in terms of \textsc{Instructions} made by CoachLM. The results, presented in Table \ref{tab:human_eval_prompt}, indicate that after the revision by CoachLM, both the \textsc{Instructions} and \textsc{Responses} received higher average scores according to all three reviewers. Notably, the improvement in \textsc{Responses} was more pronounced for the 18 samples with modified \textsc{Instructions} compared with the entire subset, implying the importance of a feasible and accurate \textsc{Instruction} in enhancing the quality of \textsc{Response}.

\subsection{Evaluation of LLM Tuned on CoachLM-revised Dataset}\label{sec:tuned_llm}
In this section, we evaluate the Alpaca-CoachLM model, which is tuned using the same settings as Alpaca \cite{alpaca}, but with the CoachLM-revised dataset replacing the \textsc{Alpaca52k} dataset. We also display our Alpaca-human model, with the human-revised subset merged into the full dataset.
\subsubsection{Compare Alpaca-CoachLM with Existing LLMs}
\paragraph{Setup} We compare our model with two groups of existing language models (LLMs). The first group is \textbf{Baseline LLMs}, which are instruction-tuned LLMs from LLaMA with the same number of parameters (\emph{i.e.}, 7B) and similar amounts of training data. To further assess the boundary of Alpaca-CoachLM, we compare it with the second group of \textbf{Stronger LLMs}. These models have larger scales (13B), are tuned with proprietary instruction datasets (e.g., LLaMA2-chat \cite{llama2}, ChatGLM2 \cite{du2022glm}), or benefit from additional feedback from RL pipelines. The four test sets used in the evaluation are described in Section \ref{sec:testset}. For each sample in a test set, PandaLM rates the candidate response against the reference responses and produces a conclusion of ``win'', ``tie'', or ``lose''. We compute three types of win rates: (1) \textbf{WR1}, which considers a ``tie'' as a half-win and is calculated as WR1=$\frac{\#win + 0.5\times \#tie}{\#all}$, where $\#all$ is the number of samples in the test set; (2) \textbf{WR2}, which excludes tied cases and is given by WR2=$\frac{\#win}{\#all - \#tie}$; and (3) \textbf{QS}, a quality score that measures the ratio of responses reaching the level of references, formulated as QS=$\frac{\#win+\#tie}{\#all}$.

\paragraph{Result} The result is shown in Table \ref{tab:pandalm_eval}. In addition to the advantage of Alpaca-human on win rates against Alpaca and Alpaca-cleaned, Alpaca-CoachLM further evolves after being trained on the fully revised dataset and outperforms all models in the baseline group, including the Vicuna-7b model \cite{vicuna2023}, which is tuned with 70k high-quality user-shared conversations with ChatGPT. Additionally, despite being smaller in scale and trained with fewer signals, Alpaca-CoachLM achieves impressive results in the group of stronger LLMs, with the highest win rates in five out of the 12 comparisons, and outperforms the 13B Vicuna model in all test sets.

\subsubsection{Human Evaluation on Alpaca-CoachLM}
In addition to automatic evaluation, human reviewers independently rated the responses generated by Alpaca-CoachLM and the original Alpaca model in the CoachLM150 test set. The reviewers were unaware of the sources of the responses. As shown in Table \ref{tab:human_eval_llm}, all reviewers consistently gave Alpaca-CoachLM a higher average score (ranging from 58.6 to 64.3) compared with the original Alpaca model. This improved performance of Alpaca-CoachLM further confirms the effectiveness of the revisions made by CoachLM, which successfully enhance the instruction-following ability of subsequently tuned LLMs by optimizing the quality of the underlying instruction dataset.

\begin{table}[htbp]
\caption{Human Evaluation on Alpaca-CoachLM and Alpaca}
\centering

\begin{tabular}{lcccc}
\toprule
\textbf{Model} & R1 & R2 & R3 & \textbf{Avg.}\\
\midrule
Alpaca & 56.6 & 58.2 & 60.9 & 58.6 \\
Alpaca-CoachLM & \textbf{61.4} & \textbf{66.9} & \textbf{64.7} & \textbf{64.3} \\
 \bottomrule
\end{tabular}
\label{tab:human_eval_llm}
\end{table}
\subsection{\colorReviewerThree{Impact of Human Input Ratio $\alpha$}}\label{sec:different_alpha}
\begin{figure}[htbp]
 \centering  
 \subfigbottomskip=-5pt 
 \subfigcapskip=-4pt \hspace{-4mm}
 \subfigure[\colorReviewerThree{Alpaca-CoachLM}]{
  \includegraphics[width=0.50\linewidth]{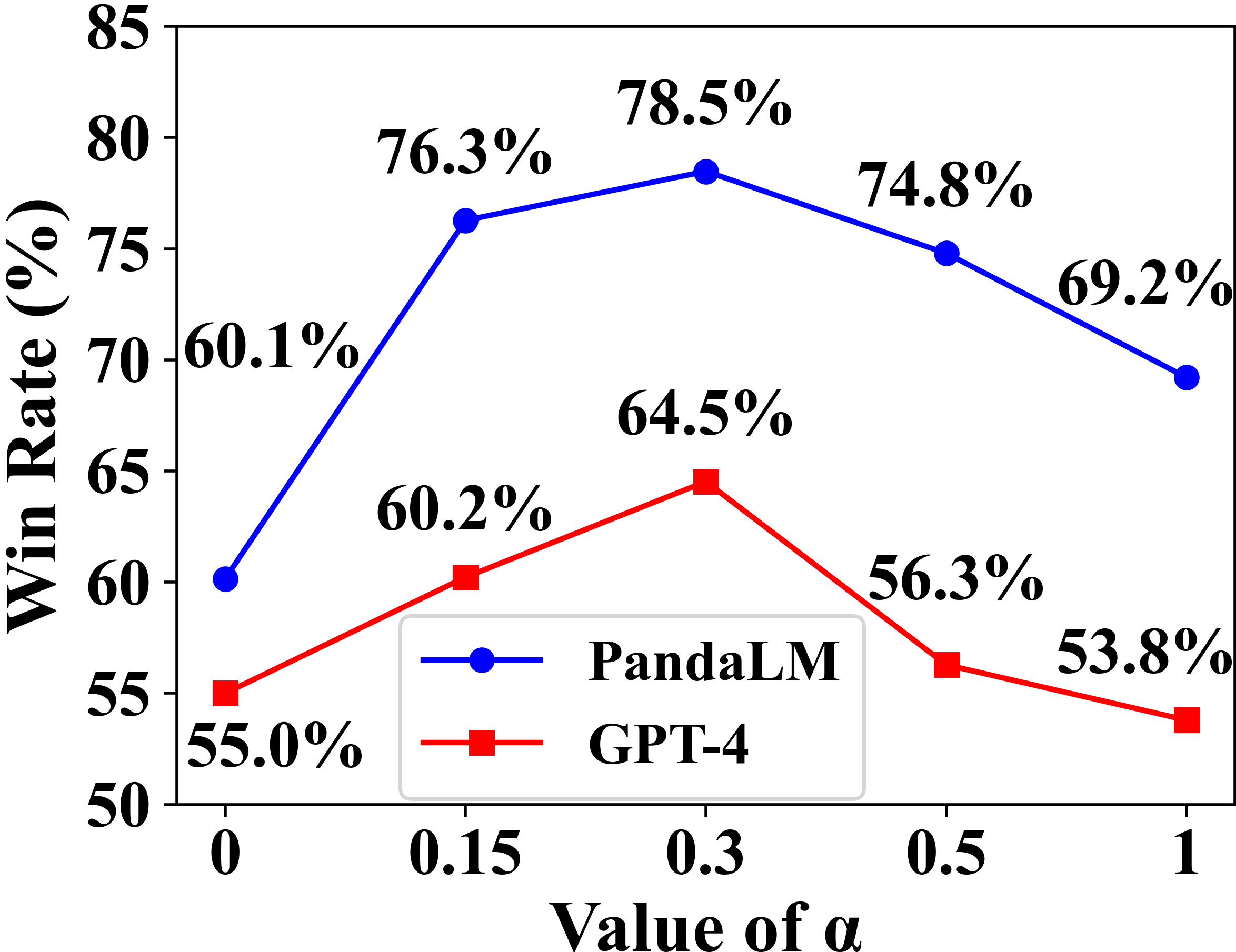}}\hspace{-1mm}
 \subfigure[\colorReviewerThree{Alpaca-human}]{
  \includegraphics[width=0.465\linewidth]{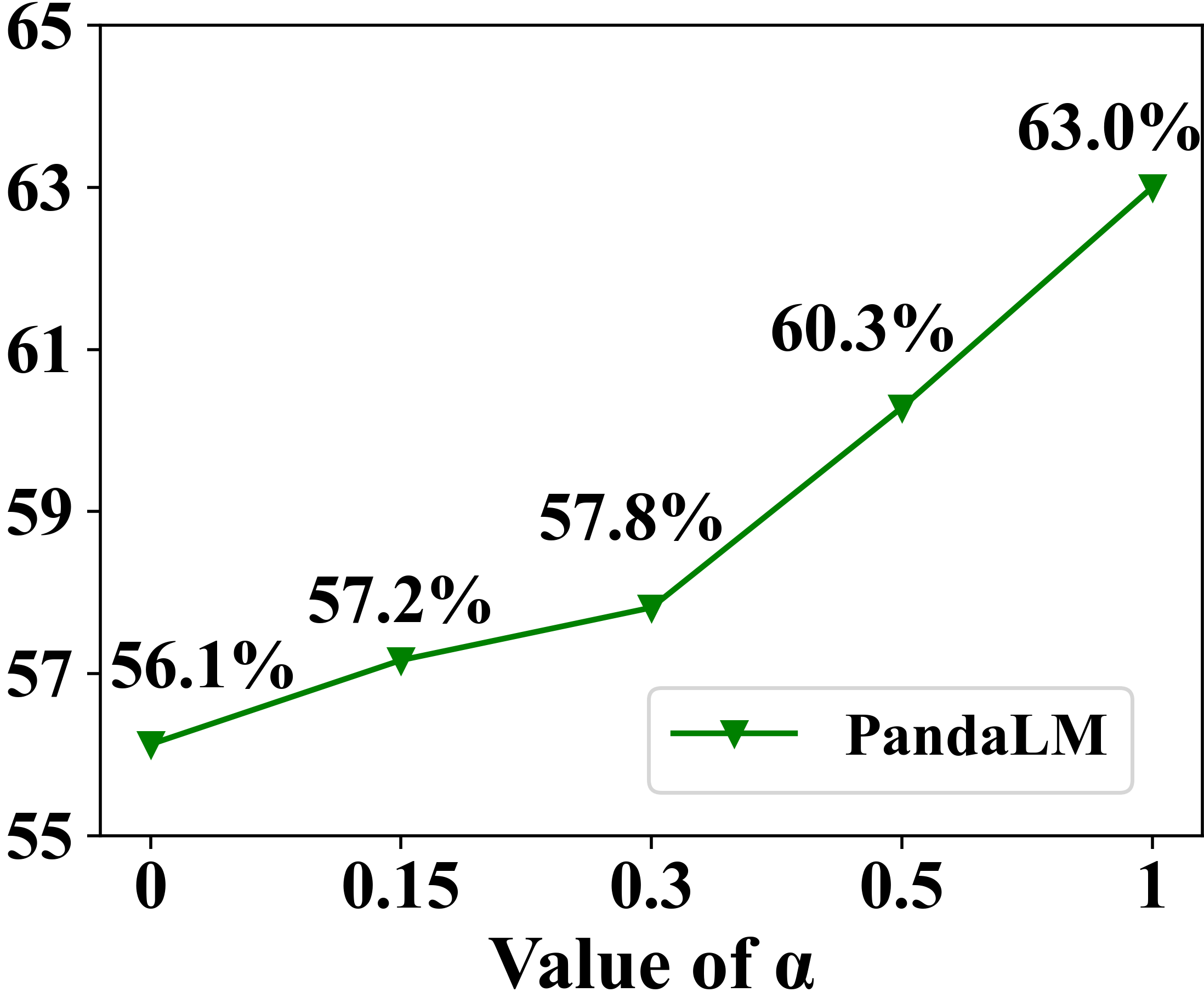}}
 \caption{\colorReviewerThree{Win rates of (a) Alpaca-CoachLM and (b) Alpaca-human against reference responses in the CoachLM150 test set with varying human input ratio $\alpha$, rated by GPT-4 and PandaLM. $\alpha$ represents ratio of human input used for training, with amount of human revision sorted from largest to smallest. $\alpha$=0 means no human input in training and $\alpha$=1 means the full human input is used. The displayed win rate is the average of WR1, WR2 and QS.}}
 \label{fig6}
\end{figure}
As is described in Section \ref{sec:alpha_design}, \colorReviewerThree{$\alpha$ determines the fraction of human input with high-quality revisions used in training. A higher $\alpha$ implies that a larger proportion of revision examples with highest edit distance is utilized. For Alpaca-CoachLM,} when $\alpha$ is set to 1, all 2.3k expert revision examples are used for CoachLM training, while a value of 0 means no training and the backbone model (ChatGLM2) is used directly for revision. By varying $\alpha$, we obtain different trained CoachLM models and subsequently tuned Alpaca-CoachLM models. Fig.~\ref{fig6}(a) shows the performance of Alpaca-CoachLM for different $\alpha$ values. Both the ratings by PandaLM and GPT-4 demonstrate a similar trend, with the highest win rate observed at $\alpha$=0.3. The win rate of Alpaca-CoachLM increases as $\alpha$ goes from 0 to 0.3, indicating the importance of high-quality expert knowledge in achieving desirable revision ability for CoachLM. However, as $\alpha$ increases beyond 0.3, the inclusion of samples with fewer modifications introduces noise in aligning CoachLM with experts, potentially lowering the quality of the CoachLM-revised dataset and decreasing the win rates of the tuned Alpaca-CoachLM. Nevertheless, the reduction in win rate caused by this noise is at most around 10\%, demonstrating the relative robustness of CoachLM.

\colorReviewerThree{Although the introduction of less-modified human input samples hindered the performance of Alpaca-CoachLM, the win rate of Alpaca-human steadily increases as more human-revised samples replace the original ones in the training dataset (Fig.~\ref{fig6}(b)). This suggests that even minor human revisions improve the quality of revised instruction pairs compared to the original counterparts, thereby enhancing the dataset used to train Alpaca-human. Based on linear fitting ($R^2=0.9799$), the win rate of Alpaca-human increases at a rate of 3.07\%/k and is estimated to surpass Alpaca-CoachLM with 7.3k human-revised samples. Notably, Alpaca-CoachLM only requires around 0.7k human-revised samples, highlighting the cost-saving advantage of CoachLM in expert labor, as it achieves the same model performance with only 9.45\% human input.}

\subsection{Different Backbone Models of CoachLM}\label{sec:different_backbone}

\begin{table}[htbp]
\caption{Performance of CoachLM with Varying Backbone Models}
\centering

\begin{tabular}{lcccc}
\toprule
\textbf{Model} & \textbf{Size} &\textbf{WR1} & \textbf{WR2} & \textbf{QS}\\
\midrule
Alpaca & - & 48.0\% & 45.7\% & 74.7\% \\
\hdashline
\noalign{\vskip 3pt}
 \multicolumn{5}{l}{\textbf{Alpaca-CoachLM (back-boned by)}} \\
 \noalign{\vskip 3pt}
LLaMA \cite{touvron2023llama} & 7B & 49.3\% & 48.6\% & 75.3\% \\
ChatGLM \cite{du2022glm} & 6B & 54.0\% & 59.1\% & 82.0\% \\
ChatGLM2 \cite{du2022glm} & 6B & \textbf{56.7\%} & \textbf{65.6\%} & \textbf{85.3\%} \\
 \bottomrule
 \noalign{\vskip 2pt}
 \multicolumn{5}{l}{Value of $\alpha$ is fixed at 1. The test set is CoachLM150.}\\
\end{tabular}
\label{tab:varying_backbone}
\end{table}

To further assess the robustness of CoachLM, we trained it with three different open-sourced backbone models: LLaMA, ChatGLM, and ChatGLM2. The win rates of the subsequently acquired Alpaca-CoachLM model on the CoachLM150 test set, evaluated by PandaLM, are displayed in Table \ref{tab:varying_backbone}. In this experiment, we kept the value of $\alpha$ fixed at 1. Our results show that Alpaca-CoachLM outperforms the original Alpaca under all backbone models, indicating the robustness of CoachLM across different backbones. Notably, we observed improved performance from LLaMA, the foundation LLM, to RL-tuned ChatGLM2, suggesting that more powerful backbones enhance the alignment ability with experts in coach instruction tuning.

\section{Discussion}\label{sec:discuss}
\subsection{CoachLM in Practice}
\begin{figure}[htbp]
    \centering
  \includegraphics[width=\linewidth]{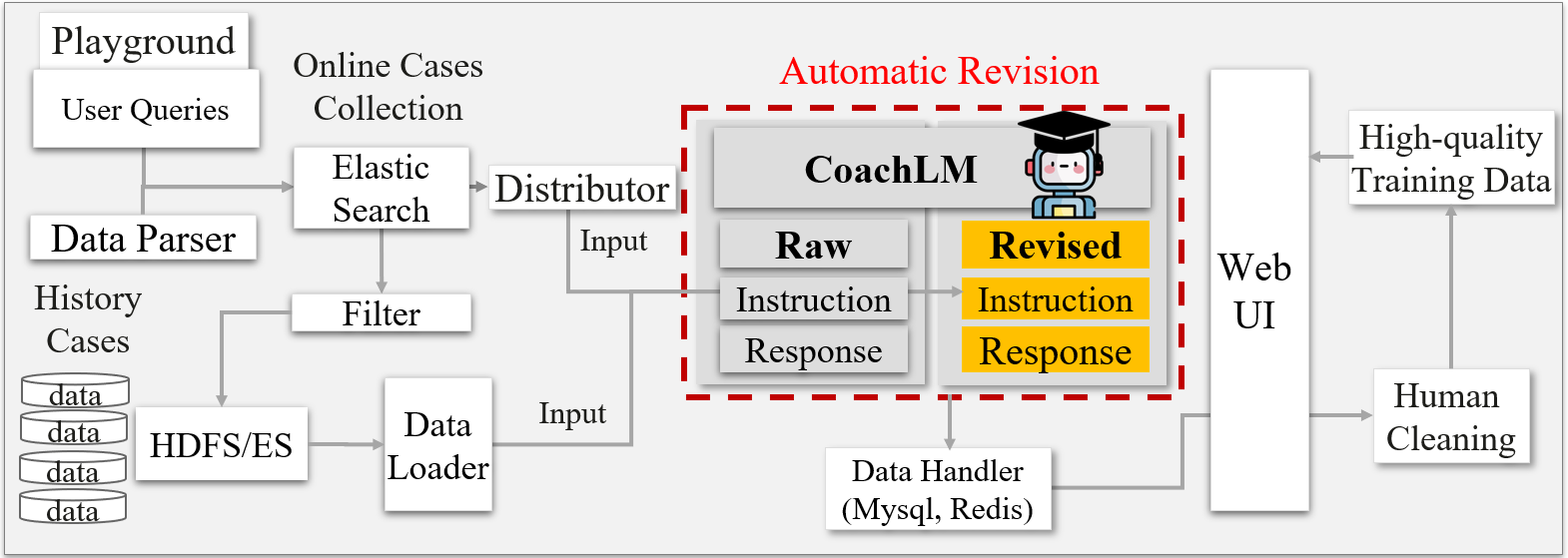}
  \caption{Architecture of an LLM data management system at Huawei integrated with CoachLM. CoachLM automatically cleans noisy instruction pairs and mitigates human workload in data cleaning. }
  \label{fig7}
\end{figure}

Given the potential advantages of CoachLM in optimizing the data collection and cleaning pipelines for LLM training, we further collaborated with a development team specializing in LLMs at Huawei and integrated CoachLM into their data management platform to facilitate LLM training. The platform, as shown in Fig.~\ref{fig7}, is responsible for gathering online user cases of deployed LLMs (with users being fully aware of their input data usage) and organizing them into high-quality training data to enable iterative enhancements of the LLMs. The data cleaning task is non-trivial, since user queries may contain noises and the responses were generated by LLMs. Previously, the platform primarily employed rule-based scripts to parse user cases into raw instruction pairs and to offer basic data filtering and cleaning. Subsequently, professional human annotators performed cleaning and revision tasks on the raw instruction pairs to curate instruction datasets of high quality for model iterations. With the incorporation of CoachLM into the pipeline, the raw instruction pairs are now subjected to automatic revisions before the manual cleaning process. Given that their human annotation guidelines also encompass dimensions such as the feasibility of instructions, as well as the correctness, richness, and helpfulness of responses, the integration of CoachLM can serve as an improved precursor for human revisions, thus mitigating the manual workload.

As of the time of writing this paper, the deployed CoachLM has successfully involved in the production of an entire batch of high-quality instruction pairs (approximately 40k). The inference process of CoachLM was executed on 1 NVIDIA A100 GPU with an inference batch size of 32, achieving an average speed of 1.19 samples per second. A comparative analysis between the current batch of data cleaning and the previous batch (with online models unchanged) reveals that the integration of CoachLM, with its revised instruction pairs serving as a precursor for human annotators, has resulted in an increase in the production efficiency of high-quality instruction pairs from around 80 per person-day to nearly 100 per person-day, while adhering to the same acceptance criteria as the previous batch. After deducting the improvement of efficiency brought by enhanced proficiency of human experts in annotation, the net improvement brought by CoachLM is estimated to be around 15-20\%, which is a significant cost saving since the inference of CoachLM on 100 samples only costs around two minutes.

\subsection{Feedbacks of CoachLM from Experts}

During the evaluation and practice of CoachLM, comments from the participating experts were actively encouraged and collected. One of the human evaluators provided feedback indicating that the responses revised by CoachLM ``generally provide more pervasive points, especially in mathematics and logical problems''. Moreover, a practitioner commented that ``CoachLM significantly augments the raw instruction pair by generating a more comprehensive structure of content, thereby enhancing the efficiency of subsequent human post-editing tasks in comparison to manual composition of the structure''.

However, there were also some concerns raised. One evaluator described a case where CoachLM did not correct the inclusion of hallucinated content but instead assumed it to be factual and further expanded upon it. Additionally, another evaluator highlighted that for certain straightforward instructions, such as determining the sum of two numbers, the level of detail in the responses revised by CoachLM may be excessive. These valuable feedbacks shed light on potential future directions for enhancing the performance of CoachLM, including refining the evaluation criteria and integrating RL signals to mitigate the occurrence of hallucinations.

\section{Related Work}\label{related_work}
\subsection{Instruction-following LLMs}
The initial investigation of the instruction-following ability of LLMs involves fine-tuning the models on a combination of multiple verbalized Natural Language Processing (NLP) datasets \cite{raffel2020exploring,wei2021finetuned}, demonstrating impressive generalization capabilities across various unseen tasks. Subsequently, instead of fine-tuning on a single task-related dataset, the mainstream LLMs have shifted towards being fine-tuned on complex human-curated instruction datasets \cite{ouyang2022training,llama2,DatabricksBlog2023DollyV2,du2022glm}. Due to the expertise requirement and high cost associated with this approach, Alpaca \cite{wang-etal-2023-self-instruct, alpaca} provides an automated method to create instruction datasets by distilling the knowledge of a teacher LLM (\emph{e.g.}, GPT-3.5). Various variants of Alpaca have been developed, including hyper-parameter optimization (Alpaca-PandaLM \cite{wang2023pandalm}), subset filtering (AlpaGasus \cite{chen2023alpagasus}), and noise cleaning (Alpaca-cleaned). Additionally, studies have explored the use of real-world user dialogue data with ChatGPT to perform instruction tuning \cite{vicuna2023,koala_blogpost_2023}.

\subsection{Data Quality in LLM}

Over the past decade, efforts have been made to improve the data quality within the AI/ML lifecycle \cite{chu2016data,schmarje2022data,sanderson2023maintaining}. When creating training datasets for LLMs, it is widely recognized that the quality of the data is more important than the quantity \cite{zhou2023lima,li2023quantity,llama2,koala_blogpost_2023,wei2023instructiongpt}. In fact, the introduction of low-quality data can harm the performance of the models. This issue is particularly pronounced in machine-generated instruction datasets, as evidenced by AlpaGasus \cite{chen2023alpagasus}, which found that out of the 52k instruction pairs in the \textsc{Alpaca52k} dataset, only 9k were of high quality. In addition to filtering-based approaches \cite{cao2023instruction,li2023quantity,chen2023alpagasus}, the Alpaca-cleaned project explored an improvement-based approach with rule-based cleaning on a small subset of the dataset.

\subsection{LLMs for Data Engineering in Industry}

\colorReviewerOne{LLM-based approaches have been increasingly utilized in various real-world data engineering tasks. For instance, Ahmed \emph{et al.} \cite{ahmed2023recommending} employed fine-tuned GPT-3.x models to facilitate cloud incident management at Microsoft. Chen \emph{et al.} \cite{chen2022software} leveraged the semantic matching capabilities of LLMs to develop a multi-vendor configuration management tool at Huawei. LLM-based programming assistants, such as Copilot \cite{GitHubCopilot2023}, have been successfully deployed in code data analysis applications, providing accurate code understanding and recommendations \cite{xu2022systematic,li2022competition,yetistiren2022assessing}. Additionally, Liu \emph{et al.} utilized LLMs to automate high-precision data analysis on tabular datasets, implementing their approach in an LCD factory and a solar cell factory \cite{liu2023jarvix}.}

\colorReviewerTwo{In comparison to existing studies, our work focuses on improving data quality in LLM training and thereby can be integrated into industrial LLM applications to improve data engineering performance. We validates the feasibility of expert-aligned revisions on instruction pairs from the entire instruction dataset. Compared with filtering-based approaches, our approach maintains the integrity of the dataset and increases the proportion of high-quality samples, thereby resulting in better performance improvements of LLMs.}

\section{Conclusion}\label{sec:conclusion}
In this study, we propose CoachLM, a novel approach to tackle the issue of unguaranteed data quality in LLM instruction tuning. Owing to the ability of automatic revisions aligned with language experts, CoachLM effectively enhances the proportion of high-quality samples in the \textsc{Alpaca52k} dataset, resulting in notable performance improvements in instruction-tuned LLMs. Additionally, the successful deployment of CoachLM in an industrial-level data management system highlights its potential advantages in the operation and maintenance lifecycle of LLMs, reducing costs associated with manual data cleaning and labeling. Future work includes training CoachLM on a larger scale of parameters, integrating RL pipelines to mitigate hallucination and validating it using a more diverse range of instruction datasets.

\clearpage
\bibliographystyle{./IEEEtran}
\bibliography{./mybib}

\end{document}